\documentclass{sig-alternate}
\newcommand{\ignore}[1]{}
\usepackage[pass]{geometry}
\usepackage{fancyhdr}
\usepackage[normalem]{ulem}
\usepackage[hyphens]{url}
\usepackage{color}
\usepackage{soul}
\usepackage{colortbl}

\usepackage[bookmarks=true,breaklinks=true,letterpaper=true,colorlinks,linkcolor=black,citecolor=blue,urlcolor=black]{hyperref}

\fancypagestyle{firstpage}{
	\fancyhf{}
	\setlength{\headheight}{50pt}
	
	\fancyhead[C]{
			} 
	\pagenumbering{arabic}
}

\title{Saving RNN Computations with a Neuron-Level Fuzzy Memoization Scheme}

\author{Franyell Silfa, Jose-Maria Arnau, Antonio Gonz\`alez\\
Computer Architecture Deparment, Universitat Politecnica de Catalunya\\
\{fsilfa, jarnau, antonio\}@ac.upc.edu
}

\begin{document}
\maketitle
\thispagestyle{firstpage}
\pagestyle{plain}

\begin{abstract}

 Recurrent Neural Networks (RNNs) are a key technology for applications
such as automatic speech recognition or machine translation. Unlike conventional
feed-forward DNNs, RNNs remember past information to improve the
accuracy of future predictions and, therefore, they are very effective
for sequence processing problems.

For each application run, recurrent layers are executed many times for processing a potentially large sequence of inputs
(words, images, audio frames, etc.). In this paper, we observe that the output
of a neuron exhibits small changes in consecutive
invocations.~We exploit this property to build a neuron-level fuzzy
memoization scheme, which dynamically caches each neuron's output and reuses it
whenever it is predicted that the current output will be similar to a previously
computed result, avoiding in this way the output computations.

The main challenge in this scheme is determining whether the new neuron's output 
for the current input in the sequence will be similar to a recently computed result. 
To this end, we extend the recurrent layer with a much simpler Bitwise Neural Network (BNN), 
and show that the BNN and RNN outputs are highly correlated: if two BNN outputs are
very similar, the corresponding outputs in the original RNN layer are likely 
to exhibit negligible changes. The BNN provides a low-cost and effective mechanism 
for deciding when fuzzy memoization can be applied with a small
impact on accuracy. 

We evaluate our memoization scheme on top of a state-of-the-art accelerator for
RNNs, for a variety of different neural networks from multiple application domains.
We show that our technique avoids more than 26.7\% of computations, 
resulting in 21\% energy savings and 1.4x speedup on average.

\end{abstract}

\section{Introduction}\label{s:introduction}

Recurrent Neuronal Networks (RNNs) represent the
state-of-the-art solution for many sequence
processing problems such as speech recognition~\cite{6638947}, machine
translation~\cite{wu2016google} or automatic caption generation~\cite{Vinyals_2015_CVPR}.
Not surprisingly, data recently published in~\cite{jouppi2017TPU} show that around
30\% of machine learning workloads in Google's datacenters 
are RNNs, whereas Convolutional Neuronal 
Networks (CNNs) only represent 5\% of the applications.
Unlike CNNs, RNNs use information
of previously processed inputs to improve
the accuracy of the output, and they
can process variable length input/output sequences.

\begin{figure*}[t!]
    \centering
    \includegraphics[width=6.5in]{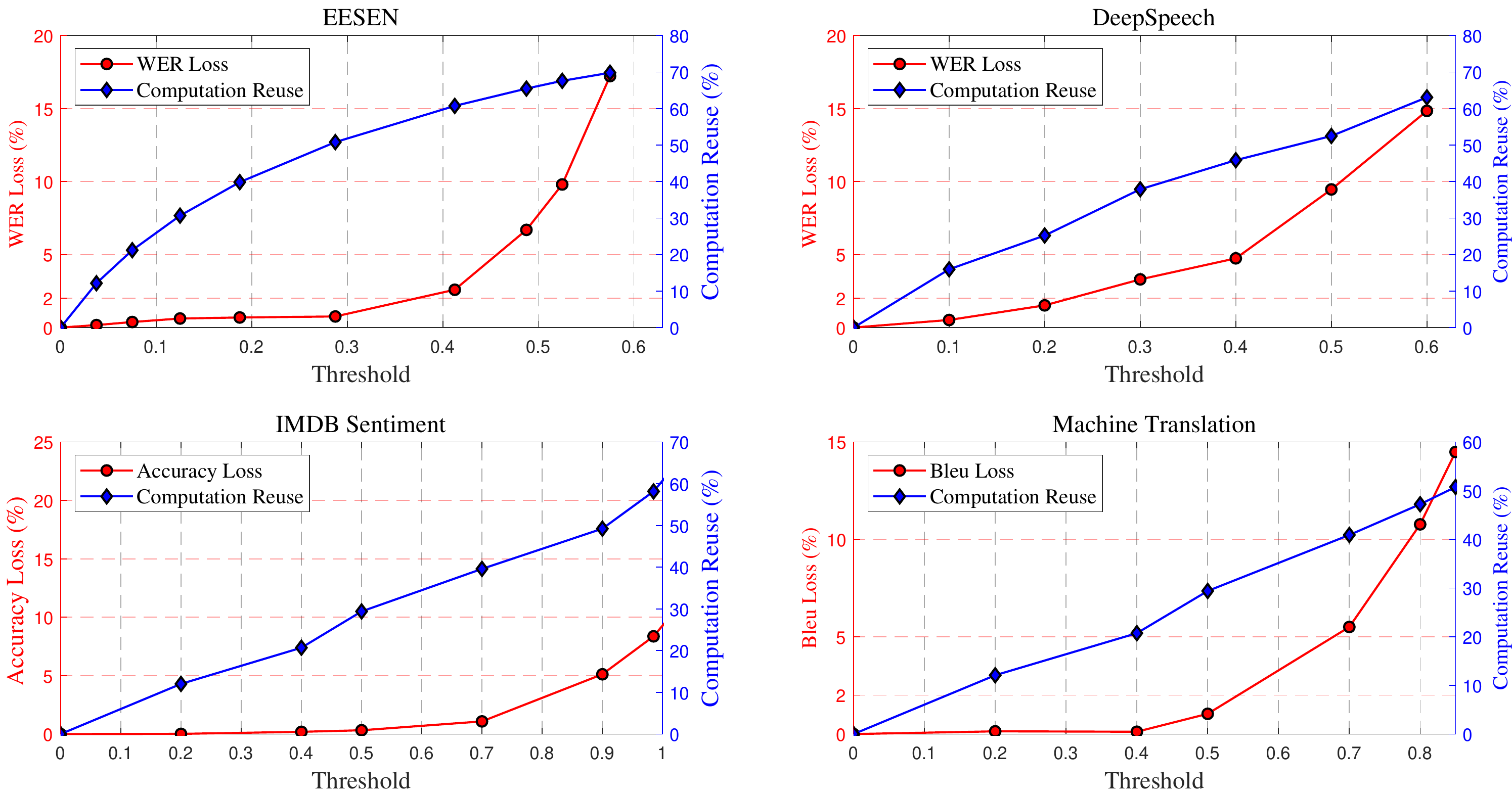}
    \caption{Accuracy loss of different RNNs versus the relative output error 
threshold using an oracle predictor. If the difference between the previous 
and current output predicted is smaller than the threshold, the memoized output 
is employed instead of calculating the new one.}
    \label{f:accuracy_th}
\end{figure*}

Although RNN training can be performed efficiently
on GPUs~\cite{appleyard2016optimizing}, RNN inference is more challenging. 
The small batch size (just one input sequence per batch)
and the data dependencies in recurrent layers severely constrain
the amount of parallelism. Hardware acceleration 
is key for achieving high-performance
and energy-efficient RNN inference and, to this end,
several RNN accelerators have been
recently proposed~\cite{Han:2017:EES:3020078.3021745,lee2016fpga,guan2017fpga,li2015fpga}.

Neurons in an RNN are recurrently executed for processing the
elements in an input sequence. An analysis of the output results reveals
that many neurons produce very similar outputs for
consecutive elements in the input sequence. On average, the relative difference 
between the current and previous output of a neuron is smaller than 23\% in our set of
RNNs, whereas previous work in~\cite{riera2018computation} has reported similar results. Since RNNs 
are inherently error tolerant~\cite{Zhang:2015:AAC:2755753.2755913},
we propose to exploit the aforementioned property to save computations by using a neuron-level 
fuzzy memoization scheme. With this approach, the outputs of a neuron
are dynamically cached in a local memoization buffer. When the next output 
is predicted to be extremely similar to the previously computed result, 
the neuron's output is read from the memoization buffer rather than 
recalculating it, avoiding all the corresponding computations and memory 
accesses.

Figure~\ref{f:accuracy_th} shows the potential benefits of this memoization
scheme by using an oracle that accurately predicts the relative difference between
the next output of the neuron and the previous output stored in the memoization
buffer. The memoized value is used when this difference is smaller
than a given threshold, shown in the x-axis of Figure~\ref{f:accuracy_th}.
As it can be seen, the RNNs can tolerate relative errors in the outputs of a
neuron in the range of 30-50\% with a negligible impact on accuracy. 
With these thresholds, a memoization scheme with an oracle predictor
can save more than 30\% of the computations.

A key challenge for our memoization scheme is how to predict
the difference between the current output and the previous output
stored in the memoization buffer, without performing all the 
corresponding neuron computations. To this end, we propose
to extend each recurrent layer with a 
Bitwise Neural Network (BNN)~\cite{kim2016bitwise}. We do this
by reducing each input and weight to one bit that represents the 
sign as described in~\cite{courbariaux2016binarized}.
We found that BNN outputs are highly correlated with
the outputs of the original recurrent layer, i.e. a similar 
BNN outputs indicates a high likelihood of having 
similar RNN output (although BNN outputs are very different to RNN outputs). The BNN is extremely small,
hardware-friendly and very effective at predicting
when memoization can be safely applied.

Note that by simply looking at the inputs, i.e. predicting
that similar inputs will produce similar outputs, might 
not be accurate. Small changes in an input that is multiplied
by a large weight will introduce a significant change
in the output of the neuron. Our BNN approach takes into account 
both the inputs and the weights.

In short, we propose a neuron-level hardware-based fuzzy
memoization scheme that works as follows. The output of 
a neuron in the last execution is dynamically cached 
in a memoization table, together with the output of the 
corresponding BNN. For every new input in the sequence,
the BNN is first computed and the result is compared with 
the BNN output stored in the memoization table. If the 
difference between the new BNN output and the cached 
output is smaller than a threshold, the neuron's cached output is used as the current output,
avoiding all the associated computations and
memory accesses in the RNN. Otherwise, the neuron is
evaluated and the memoization table is updated.

Note that only using the BNN would result in a large
accuracy loss as reported elsewhere~\cite{rastegari2016xnor}. In this paper, we take a completely
different approach and use the BNN to predict when
memoization can be safely applied with negligible impact
on accuracy. The inexpensive BNN is computed for every
element of the sequence and every neuron, whereas the 
large RNN is evaluated on demand as indicated by the BNN.
By doing so, we maintain high accuracy while saving
more than 26.7\% of RNN computations.

In this paper, we make the following contributions:
\begin{itemize}
\item We provide an evaluation of the outputs of neurons in recurrent layers,
and show that they exhibit small changes in consecutive executions.
\item We propose a fuzzy memoization scheme that
avoids more than 26.7\% of neuron evaluations by reusing previously computed
results stored in a memoization buffer.
\item We propose the use of a BNN to determine when memoization
can be applied with small impact on accuracy. We show that BNN
and RNN outputs are highly correlated.
\item We show that the BNN predictor's accuracy improves significantly when it is also included during the training.
\item We implement our neuron-level memoization scheme on top of a 
state-of-the-art RNN accelerator. The required hardware introduces
a negligible area overhead, while it provides 1.4x speedup and
21\% energy savings on average for several RNNs.
\end{itemize}

\section{Background}\label{s:background}

\subsection{Recurrent Neural Networks}\label{s:rnn_networks}

A Recurrent Neural Network (RNN) is a state-of-the-art machine learning approach that 
has achieved tremendous success in applications such as machine translation or video
description. The key characteristic of RNNs is that they include 
loops, a.k.a. recurrent connections, that allow the information to persist from one time-step 
of execution to the next ones and, hence, they have the potential to use unbounded 
context information (i.e. past or future) to make predictions. Another important 
feature is that RNNs are recurrently executed for every element of the 
input sequence and, thus, they are able to handle input and output with 
variable length. Because of these characteristics, RNNs provide an effective 
framework for sequence-to-sequence applications (e.g. machine translation), where they
outperform feed forward Deep Neural Networks (DNNs)~\cite{greff2016lstm, schuster1997bidirectional}.

\begin{figure}[t!]
	\centering
	\includegraphics[width=3.375in]{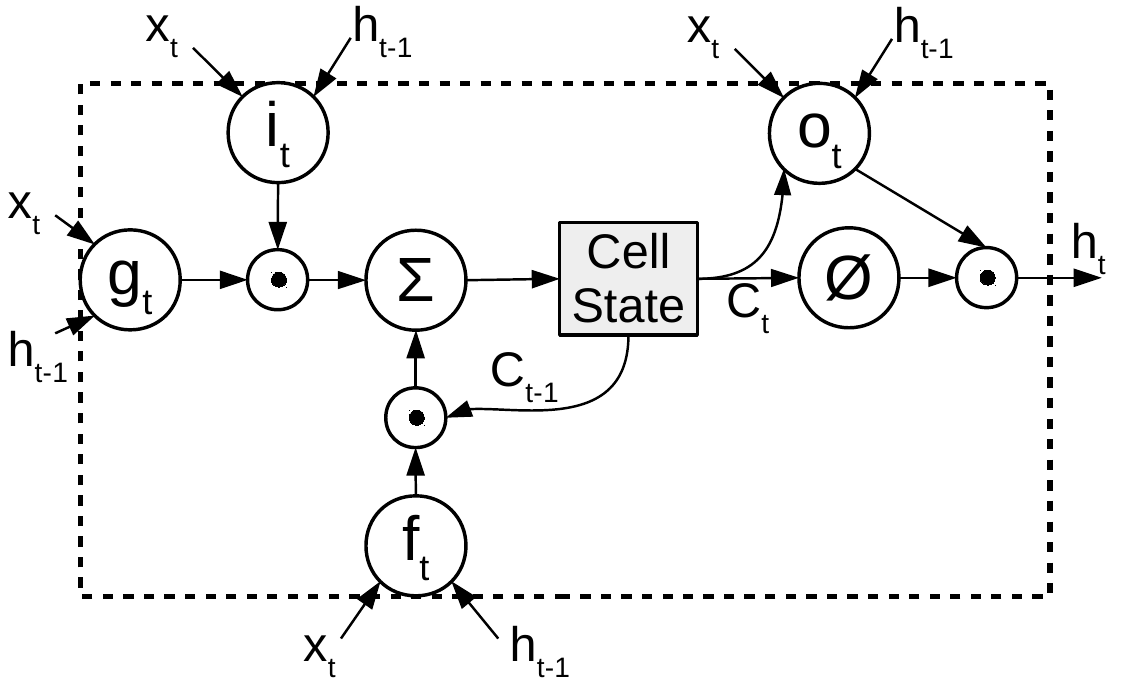}
	\caption{Structure of a LSTM cell. $\odot$ denotes an element-wise 
		multiplication of two vectors. $\phi$ denotes the hyperbolic
		tangent.}
	\label{f:lstm_cell}
\end{figure}

\begin{figure}[t!]
	\centering
	\includegraphics[width=3.375in]{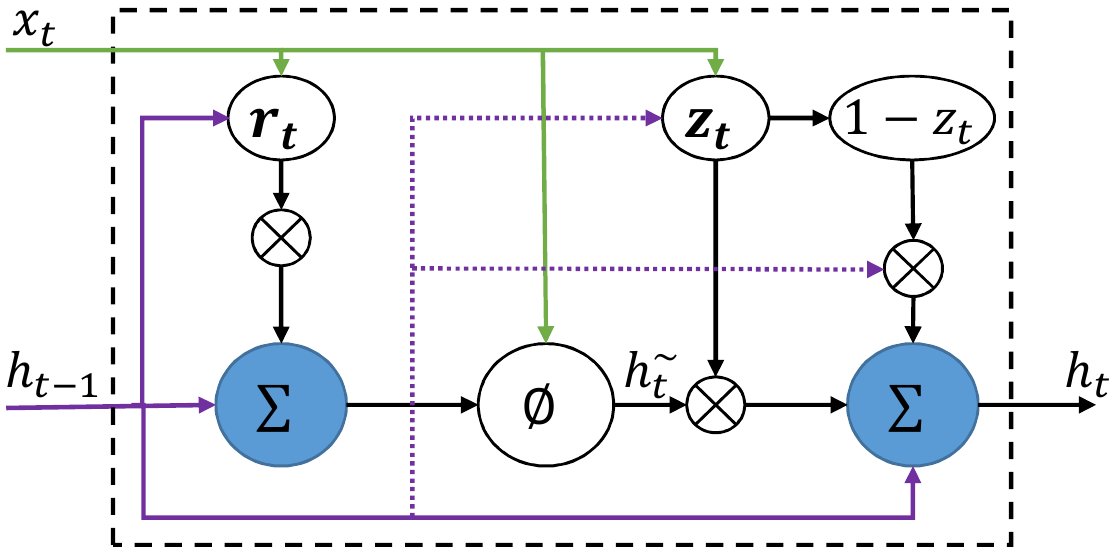}
	\caption{Structure of a GRU cell.}
	\label{f:gru_cell}
\end{figure}

Basic RNN architectures can capture and exploit short term dependencies
in the input sequence. However, capturing long term dependencies is challenging 
since useful information tend to dilute over time. In order 
to exploit long term dependencies, Long Short Term Memory (LSTM)~\cite{hochreiter1997long} and Gated Recurrent Units (GRU)~\cite{cho14Gru} networks 
were proposed. These types of RNNs represent the most successful and widely used RNN architectures. They 
have achieved tremendous succcess for a variety of applications such as speech
recognition~\cite{miao2015eesen, deepspeech2}, machine translation~\cite{britzGLL17} and video 
description~\cite{Vinyals_2015_CVPR}. The next subsections provide further details on 
the structure and behavior of these networks.

\subsubsection{Deep RNNs}\label{s:lstm_rnn}

RNNs are composed of multiple layers that are stacked together to create deep RNNs. Each of these layers consists of an LSTM or a GRU cell. In addition, these layers can be unidirectional or bidirectional. Unidirectional layers only use past  information to make predictions, whereas bidirectional LSTM or GRU networks use both past and future context.

The input sequence ($X$) is composed of $N$ elements, i.e. $X = [x_1, x_2, ..., x_N]$, 
which are processed by an LSTM or GRU cell in the forward direction, i.e. from $x_1$ to $x_N$.
For backward layers in bidirectional RNNs, the input sequence is evaluated in the backward direction, 
i.e from $x_N$ to $x_1$.

\subsubsection{LSTM Cell}\label{s:lstm_cell}

Figure~\ref{f:lstm_cell} shows the structure of an LSTM cell. The key component is the 
cell state ($c_t$), which is stored in the cell memory. The cell state is updated
by using three fully connected single-layer neural networks, a.k.a. gates. The input gate, 
($i_t$, whose computations are shown in Equation~\ref{e:input_gate}) decides how much 
of the input information, $x_t$, will be added to the cell state. The forget gate ($f_t$,
shown in Equation~\ref{e:forget_gate}) determines how much information will be erased
from the cell state ($c_{t-1}$). The updater gate ($g_t$, Equation~\ref{e:update_gate}) 
controls the amount of input information that is being considered a candidate to update the cell state ($c_t$).
Once these three gates are executed, the cell state is updated according to
Equation~\ref{e:cell_state}. Finally, the output gate ($o_t$, Equation~\ref{e:output_gate}) 
decides the amount of information that will be emitted from the cell to create the output ($h_t$).

Figure~\ref{f:lstm_equations} shows the computations carried out by an LSTM cell. 
As it can be seen, a neuron in each gate has two types of connections: forward
connections that operate on $x_t$ and recurrent connections that take as input
$h_{t-1}$. The evaluation of a neuron in one of these gates requires a dot
product between weights in forward connections and $x_t$, and another
dot product between weights in recurrent connections and $h_{t-1}$. Next, a 
peephole connection~\cite{gers2000recurrent} and a bias are also applied, 
followed by the computation of an activation function, typically
a sigmoid or hyperbolic tangent.

\subsubsection{GRU Cell}\label{s:gru_cell}
 Analogous to an LSTM cell, a GRU cell includes gates to control the flow of information inside the cell. However, GRU cells
 do not have an independent memory cell (i.e. cell state). As it can be seen in Figure~\ref{f:gru_cell}, in a GRU cell the update gate ($z_t$) controls how much of the candidate information ($g_t$) is used
 to update the cell activation. On the other hand, the reset gate ($r_t$) modulates the amount of information that is removed from the previous computed state. Note that GRUs do not include an output gate and, hence, the whole state of 
 the cell is exposed at each timestep. The computations carried out by each gate in a GRU cell are very similar to those in Equations~\ref{e:input_gate}, ~\ref{e:forget_gate} and ~\ref{e:update_gate}. We omit them for the sake of brevity, the exact details are provided in~\cite{cho14Gru}. For the rest of the paper, we used the term RNN cell to refer to both LSTM and GRU cells.

\begin{figure}[t!]
	\centering
	\begin{align}
	i_t = \sigma(W_{ix} x_t + W_{ih} h_{t-1}  + b_i)
	\label{e:input_gate}
	\end{align}
	\begin{align}
	f_t = \sigma(W_{fx} x_t + W_{fh} h_{t-1}  + b_f)
	\label{e:forget_gate}
	\end{align}
	\begin{align}
	g_t = \phi(W_{gx} x_t + W_{gh} h_{t-1} + b_g)
	\label{e:update_gate}
	\end{align}
	\begin{align}
	c_t = f_t \odot c_{t-1} + i_t \odot g_t
	\label{e:cell_state}
	\end{align}
	\begin{align}
	o_t = \sigma(W_{ox} x_t + W_{oh} h_{t-1}  + b_o)
	\label{e:output_gate}
	\end{align}
	\begin{align}
	h_t = o_t \odot \phi(c_t)
	\label{e:cell_output}
	\end{align}
	\caption{Computations of an LSTM cell. $\odot$, $\phi$, and $\sigma$ denote element-wise multiplication,
		hyperbolic tangent and sigmoid function respectively.}
	\label{f:lstm_equations}
\end{figure}

\subsection{Binarized Neural Networks}\label{s:bnn_networks}

State-of-the-art DNNs typically consist of millions of parameters (a.k.a. weights) 
represented as floating point numbers using 32 or 16 bits and, hence, their storage 
requirements are quite large. Linear quantization may be used to reduce memory
footprint and improve performance \cite{wu2016google, jouppi2017TPU}. In addition, 
real-time evaluation of DNNs requires a high energy cost. As an attempt to improve
the energy-efficiency of DNNs, Binarized Neural Networks (BNNs)~\cite{courbariaux2016binarized} or
Bitwise Neural Networks~\cite{kim2016bitwise} are a promising alternative to conventional DNNs. BNNs use one-bit weights and inputs that are constrained to +1 or -1. Typically, the binarization is done using the following function:

\begin{equation}
x^b = \begin{cases}
+1    &\text{if }  x>=0, \\
-1         &\text{otherwise, }  
\end{cases}
\label{e:x_binarization}
\end{equation}
where $x$ is either a weight or an input and $x^b$ is the binarized value which is stored
as 0 or 1. Regarding the output of a given neuron, its computation is analogous to
conventional DNNs, but employing the binarized version of weights and inputs, 
as shown in Equation~\ref{e:binarized_neuron}:
\begin{equation}
y_t^b = \sum w^bx_t^b
\label{e:binarized_neuron}
\end{equation}
where $w^b$ and $x^b_t$ are the binarized weight and input vectors respectively. Note that
evaluating the neuron output ($y_t^b$) only involves multiplications and additions that,
with binarized operands, can be computed with XNORs and integer adders. BNN evaluation 
is orders of magnitude more efficient, in terms of both performance and energy, than conventional
DNNs~\cite{courbariaux2016binarized}. Nonetheless, DNNs and RNNs still deliver significantly higher 
accuracy than BNNs~\cite{rastegari2016xnor}. 

\subsection{Fuzzy Memoization}\label{s:lfuzzy_memoization}

Memoization is a well-known optimization technique used to improve performance 
and energy consumption that has been used both in software~\cite{acar2003SM} and hardware~\cite{gonzalezTM99}. In some applications, a given function is executed many times, but the inputs of different executions are not always different. Memoization exploits this fact to avoid these redundant computations by reusing the result of a previous evaluation. In general, the 
first time an input is evaluated, the result is cached in a memoization table. Subsequent
evaluations probe the memoization table and reuse previously cached results if the current input matches a previous execution.

In a classical memoization scheme, a memoized value is only reused when it is known to be
equal to the real output of the computation. However, for some applications such as 
multimedia~\cite{alvarez2005FMF}, graphics~\cite{arnau2014}, and 
neural networks~\cite{Zhang:2015:AAC:2755753.2755913}, this scheme can be extended to tolerate a small loss in accuracy with negligible impact in the quality of the results, and is normally referred to as fuzzy memoization.

\section{Neuron Level Memoization}\label{s:technique}

In this section, we propose a novel memoization scheme to reduce computations
and memory accesses in RNNs. First, we discuss the main
performance and energy bottlenecks on state-of-the-art hardware accelerators 
for RNN inference. Next, we introduce the key idea for our 
neuron-level fuzzy memoization technique. Finally, we describe the hardware 
implementation of our technique.

\begin{figure}[t!]
	\centering
	\includegraphics[width=3.375in]{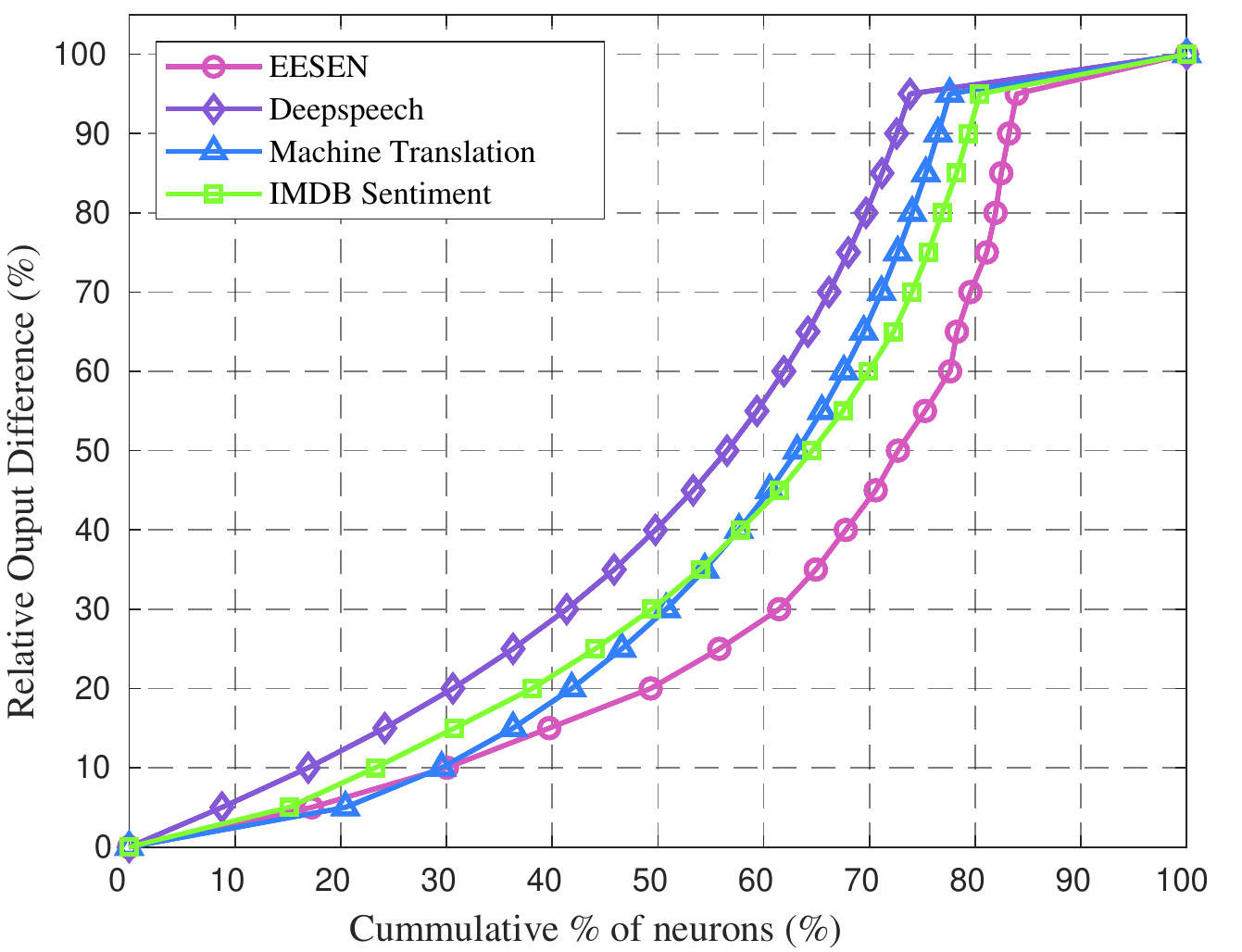}
	\caption{Relative change in neuron output between consecutive input elements. }
	\label{f:output_change}
\end{figure}

\subsection{Motivation}

As shown in Figure \ref{f:lstm_equations}, RNN inference involves the evaluation 
of multiple single-layer feed-forward neural networks or gates that, from a computational 
point of view, consist of multiplying a weight matrix by an input vector ($x_t$ for
forward connections and $h_{t-1}$ for recurrent connections). Typically, 
the number of elements in the weight matrices ranges from a few thousands to millions of 
elements and, thus, fetching them from on-chip buffers or main memory is one of 
the major sources of energy consumption. Not surprisingly, it accounts for up to 
80\% of the total energy consumption in state-of-the-art accelerators~\cite{silfa2017epur}.
For this reason, a very effective way of saving energy in RNNs is to avoid 
fetching the synaptic weights. In addition, avoiding the corresponding computations also increases
the energy savings. In this work, we leverage fuzzy memoization to selectively avoid neurons 
evaluations and, hence, to avoid their corresponding memory 
accesses and computations. For fuzzy memoization
to be effective, applications must be tolerant to small errors and its hardware implementation
must be simple. In the next sections, we show that RNNs are resilient to 
small errors in the outputs of the neurons, and we provide an efficient
implementation of the memoization scheme that requires simple hardware support.
\begin{figure}[t!]
	\begin{equation}
	\delta =\left| \frac   {y_t^o - y_{m}}{y^o_{t}} \right|
	\label{e:delta_error}
	\end{equation}
	\begin{equation}
	y_t = \begin{cases}
	y_{m}    &\text{if }  \delta<=\theta \\
	y_t^o         &\text{otherwise, } 
	\end{cases}
	\label{e:yt_basic}
	\end{equation}
	\begin{equation}
	y_m = \begin{cases}
	y^o_{t}         &\text{if }  \delta>\theta \\
	\text{not updated}   &\text{otherwise, } 
	\end{cases}
	\label{e:ym_basic}
	\end{equation}
	\caption{Neuron Level memoization with Oracle Predictor. $y_t$ is the neuron output. $y_m$ corresponds to the memoized evaluation and $y_t^o$ is the output of the Oracle predictor. $\delta$, $\theta$ are the relative error and the maximum allowed output error respectively.}
	\label{f:oracle_memoization}	
\end{figure}

\subsubsection{RNNs Redundancy}\label{s:rnn_redundancy}

Memoization schemes rely on a high degree of redundancy in the computations. For RNNs, a key 
observation is that the output of a given neuron tends to change lightly between consecutive 
input elements. Note that RNNs are used in sequence processing problems such as speech recognition
or video processing, where RNN inputs in consecutive time-steps tend to be extremely similar. Prior work in~\cite{riera2018computation} reports high similarity across consecutive frames of audio or video. Not surprisingly,
our numbers for our set of RNNs also support this claim. Figure~\ref{f:output_change} shows the relative difference between consecutive outputs of a neuron in our set of RNNs. As it can be seen, a neuron's output exhibits minor changes (less than 10\%)
for 25\% of consecutive input elements. On average, consecutive outputs change by 23\%. Furthermore, RNNs can 
tolerate small errors in the neuron output~\cite{Zhang:2015:AAC:2755753.2755913}. This observation 
is supported by data shown in Figure~\ref{f:accuracy_th}, where the accuracy curve shows the accuracy 
loss when the output of a neuron is reused using fuzzy memoization, for different thresholds
(x-axis) that control the aggressiveness of the memoization scheme. For this study, 
the relative error ($\delta$) between a predicted neuron output ($y_t^p$) and a previously 
cached neuron output ($y_m$) is used as the discriminating factor to decide whether the previous 
output is reused, as shown in Figure~\ref{f:oracle_memoization}. To evaluate the potential
benefits of a memoization scheme, the predicted value is provided by an Oracle predictor, 
which is 100\% accurate, i.e., its prediction is always equal to the neuron output ($y_t^p = y_t$).
As shown in Figure~\ref{f:accuracy_th}, neurons can
 tolerate a relative output error between 0.3 and 0.5 without significantly affecting 
the overall network accuracy (i.e., accuracy loss smaller than 1\%). On the other hand, the reuse curve shows the 
percentage of neuron computations that could be avoided through this memoization with an 
Oracle predictor. Note that by allowing neurons to have an output error between 0.3 to 0.5, 
at least 30\% of the total network computations could be avoided.

The memoization scheme must add a small overhead to the system to achieve significant savings. Therefore, the critical challenge is approximating the Oracle predictor's behavior with simple hardware to decide when memoization can be safely applied with a negligible impact on the overall RNN accuracy. We describe an effective solution in the next section.

\subsubsection{Binary Network Correlation}\label{s:bnn_correlation}

A key challenge for an effective fuzzy memoization scheme is to identify when the next 
neuron output will be similar to a previously computed (and cached) output. Note that
having similar inputs does not necessarily result in similar outputs, as inputs with
small changes might be multiplied by large weights. Our proposed approach is based on 
a Bitwise Neural Network (BNN). In particular, each fully-connected neural 
network (NN) is extended to an equivalent BNN, as described in 
Section~\ref{s:memoizing_overview}. We use BNNs for two reasons. 
First, the outputs of a BNN and its corresponding original NN are highly 
correlated~\cite{andersonB17}, i.e., a small change in a BNN output indicates that the neuron's 
output in the original NN is likely to be similar. 
Second, BNNs can be implemented with extremely low hardware cost.

\begin{figure}[t!]
	\centering
	\includegraphics[width=3.375in]{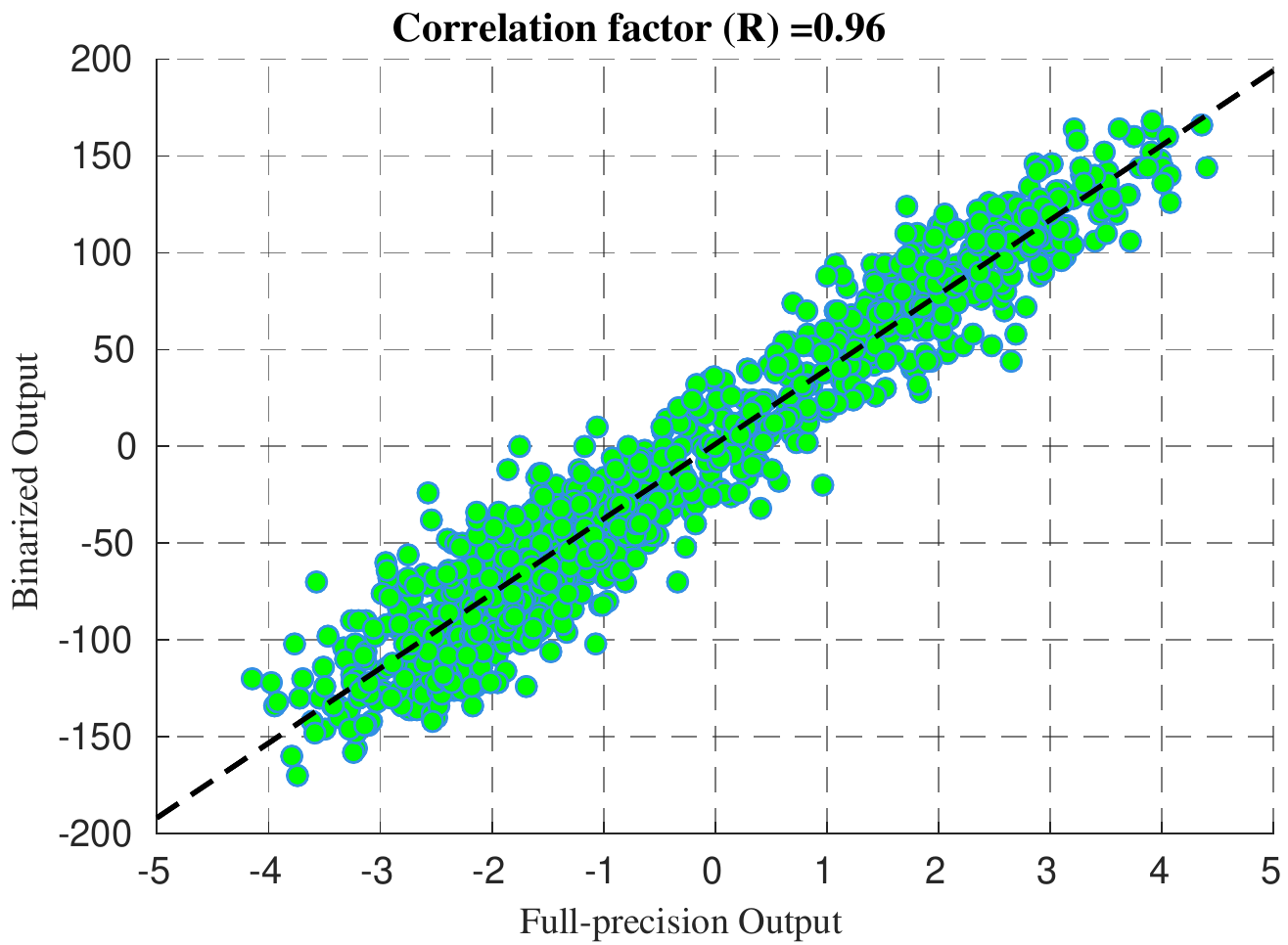}
	\caption{Outputs of the binarized neurons (y-axis) versus outputs of the full-precision neurons (x-axis) in EESEN: an RNN for speech recognition. BNN and RNN outputs are highly correlated, showing a correlation coefficient of 0.96.}
	\label{f:output_correlation_EESEN}
\end{figure}

\begin{figure}[t!]
	\centering
	\includegraphics[width=3.375in]{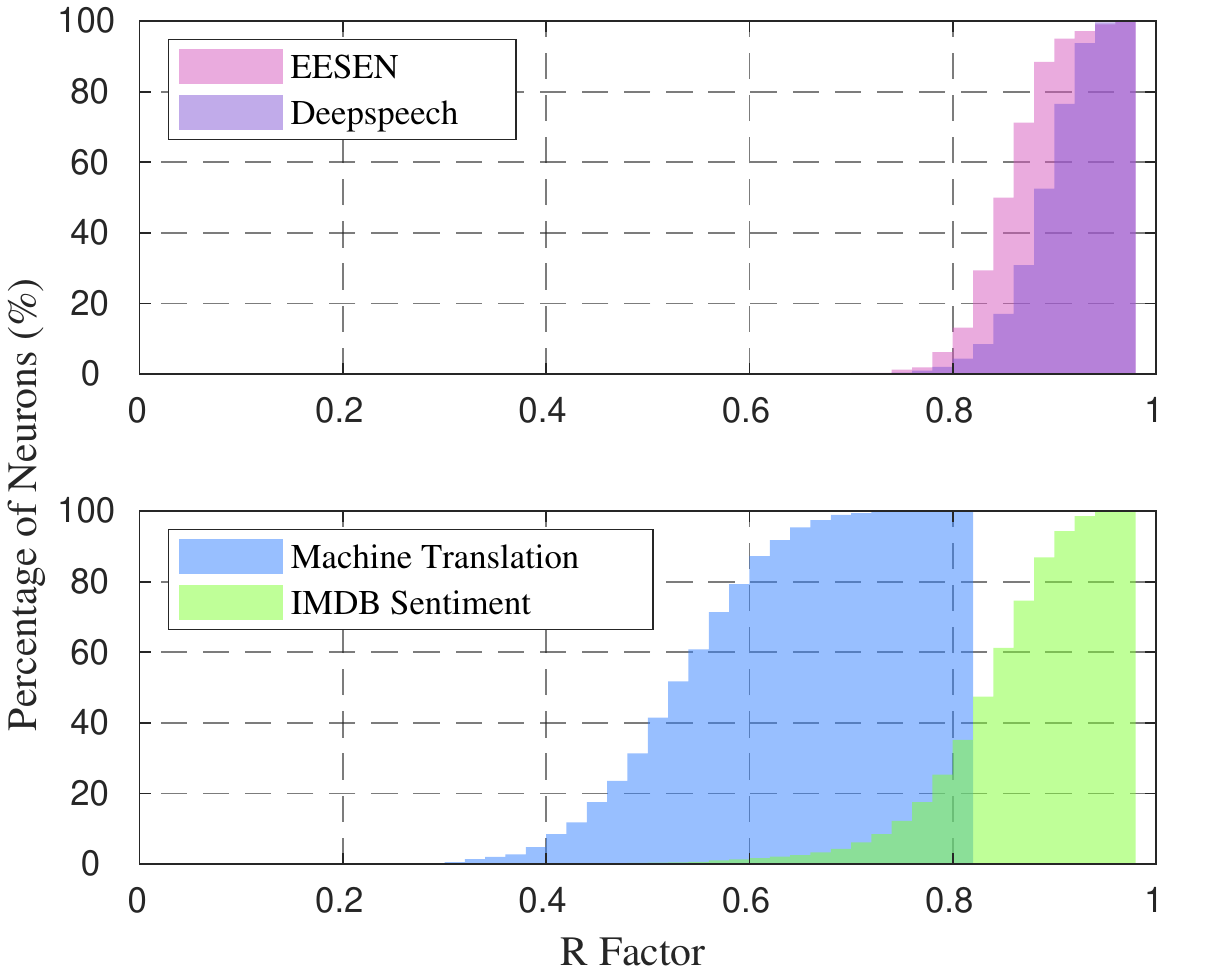}
	\caption{Correlation factor between the neuron output computed using full precision and the output computed with a BNN. }
	\label{f:output_correlation}
\end{figure}

Regarding the correlation between BNN and RNN, Anderson et al.~\cite{andersonB17} show that the binarization
approximately preserves the dot-products that a neural network performs for computations. Therefore, there should be a
high correlation between the outputs of the full-precision neuron and the outputs of the 
corresponding binarized neuron. We have empirically validated the dot product preservation property for our 
set of RNNs. Figure~\ref{f:output_correlation_EESEN} shows the linear correlation between
RNN outputs and the corresponding BNN outputs for EESEN network. Although the range of the outputs of
the full-precision (RNN) and binarized (BNN) dot products are significantly different,
their values exhibit a strong linear correlation (correlation coefficient of 0.96). On the other
hand, Figure~\ref{f:output_correlation} shows the histogram of the correlation coefficients
for the neurons in four different RNNs. As it can be seen, correlation between binarized
and full-precision neurons tend to be high for all the RNNs. More specifically, for the networks EESEN, IMDB SENTIMENT, and DEEPSPEECH, 85\% of the 
neurons have a linear correlation factor greater than 0.8 and for the Machine Translation network most of them have a correlation factor greater than 0.5.~These results indicate that if
the output of a binarized neuron shows very small changes with respect to a previously
computed output, it is very likely that the full-precision neuron will also show small
changes and, hence, memoization can be safely applied.

As shown in Equation~\ref{e:binarized_neuron}, 
the output of a given neuron in a BNN can be computed with an N-bit XOR operation 
for bit multiplication and an integer adder to sum the resulting bits. These 
two operations are orders of magnitude cheaper than those required by the traditional 
data representation (i.e., FP16). Therefore, a BNN represents a low overhead and 
accurate manner to infer when the output of a neuron is likely to exhibit significant 
changes with respect to its recently computed outputs.
  
\subsection{Overview}\label{s:memoizing_overview}

\begin{figure}[t!]
	\centering
	\includegraphics[width=3.375in]{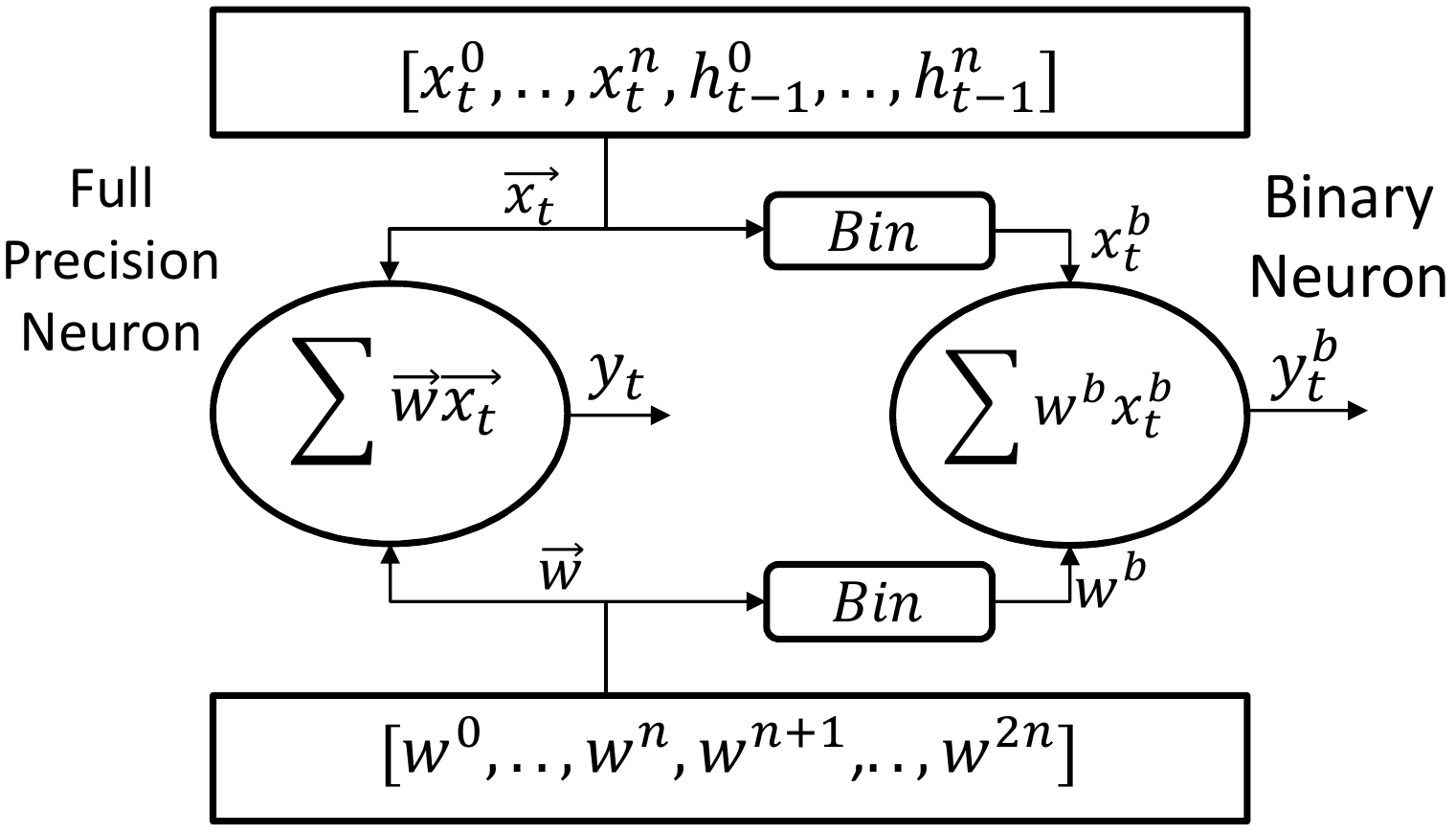}
	\caption{The figure illustrates how a binary neuron is created from a full-precision neuron in
		the RNN network. \textit{Bin} is the binarization function shown in Equation \ref{e:x_binarization}.
		Peepholes, bias and activation functions are omitted for simplicity.}
	\label{f:lstm_binary_mapping}
\end{figure}

\begin{figure}[t!]
	\begin{equation}
	\epsilon_t^b =\left| \frac   {y_t^b - y^b_{m}}{y^b_{t}} \right|
	\label{e:bin_epsilon}
	\end{equation}
	
	\begin{equation}
	\delta_t^b =\sum_{i=m}^{i=t} \epsilon_i^b 
	\label{e:bin_delta_error}
	\end{equation}
	\begin{equation}
	y_t = \begin{cases}
	y_{m}    &\text{if }  \delta_t^b<=\theta \\
	\text{evaluate neuron}        &\text{otherwise, }  
	\end{cases}
	\label{e:yt_bin}
	\end{equation}
	\begin{equation}
	y_m = \begin{cases}
	y_{t}    &\text{if }  \delta_t^b>\theta \\
	\text{not updated}         &\text{otherwise,}  
	\end{cases}
	\label{e:ym_bin}
	\end{equation}
	\begin{equation}
	y_m^b = \begin{cases}
	y^b_t    &\text{if }  \delta_t^b>\theta \\
	\text{not updated}      &\text{otherwise,}
	\end{cases}
	\label{e:vm_bin}
	\end{equation}
	\begin{equation}
	\delta_t^b = \begin{cases}
	0.0   &\text{if }  \delta_t^b>\theta \\
	\text{not updated}      &\text{otherwise,}
	\end{cases}
	\label{e:delta_bin_reset}
	\end{equation}
	\caption{Neuron level fuzzy memoization with binary network as predictor. $y_t$, $y_m$ 
		correspond to the neuron current and memoized output computed by the LSTM Network. $y^b_t$, $y^b_m$ 
		are the current and memoized output computed by the Binary Network. $\epsilon_t^b$ is the
		relative difference between BNN outputs. $\delta_t^b$ is the summation of relative differences
		in successive time-steps.}
	\label{f:binary_memo_equations}
\end{figure}

The target of our memoization scheme is to reuse a recently computed neuron output, 
$y_{m}$, as the output for the current time-step, $y_t$, provided that they are very 
similar. Reusing the cached neuron output avoids performing all the corresponding 
computations and memory accesses. To determine whether $y_t$ will be similar to 
$y_{m}$, we use a BNN as a predictor.

In our memoization scheme, we extend the RNN with a much simpler BNN.
The BNN model is created by mirroring the full precision trained model
of an LSTM or GRU gate, as illustrated in Figure~\ref{f:lstm_binary_mapping}.
More specifically, each neuron is binarized by applying the binarization 
function shown in Equation~\ref{e:x_binarization} to its corresponding 
set of weights. Therefore, in an gate, every neuron $n$ with
weights vector $\vec{w}$ is mirrored to a neuron $n^{b}$ with
weights vector $\vec{w}^{b}$ corresponding to the element-wise 
binarization of $\vec{w}$.

Our scheme stores recently computed outputs for the binary neuron
$n^{b}$ and its associated full-precision neuron $n$. We refer to
these memoized values as $y_m^b$ and $y_m$, respectively.
On every time-step $t$, the binarized version of the neuron,
$n^{b}$, is evaluated first obtaining $y_t^b$. Next, we compute
the relative difference, $\epsilon_t^b$, between $y_t^b$ and
$y_m^b$, i.e. the current and memoized outputs of the BNN,
as shown in Equation~\ref{e:bin_epsilon}. 
If $\epsilon_t^b$ is small, i.e., if the BNN outputs
are similar, it means that the outputs
of the full precision neuron are likely to be similar.
As we discuss in Section~\ref{s:bnn_correlation},
there is a high correlation between BNN and RNN
outputs. In this case, we can reuse the memoized output $y_m$ as the
output of neuron $n$ for the current time-step, avoiding
all the corresponding computations. If the
relative difference $\epsilon_t^b$ is significant, we compute 
the full-precision neuron output, $y_t$, and update our memoization buffer,
as shown in Equations~\ref{e:ym_bin}, \ref{e:vm_bin} and \ref{e:delta_bin_reset}
so that these values can be reused in subsequent time-steps.

\begin{figure}[t!]
	\centering
	\includegraphics[width=3.375in]{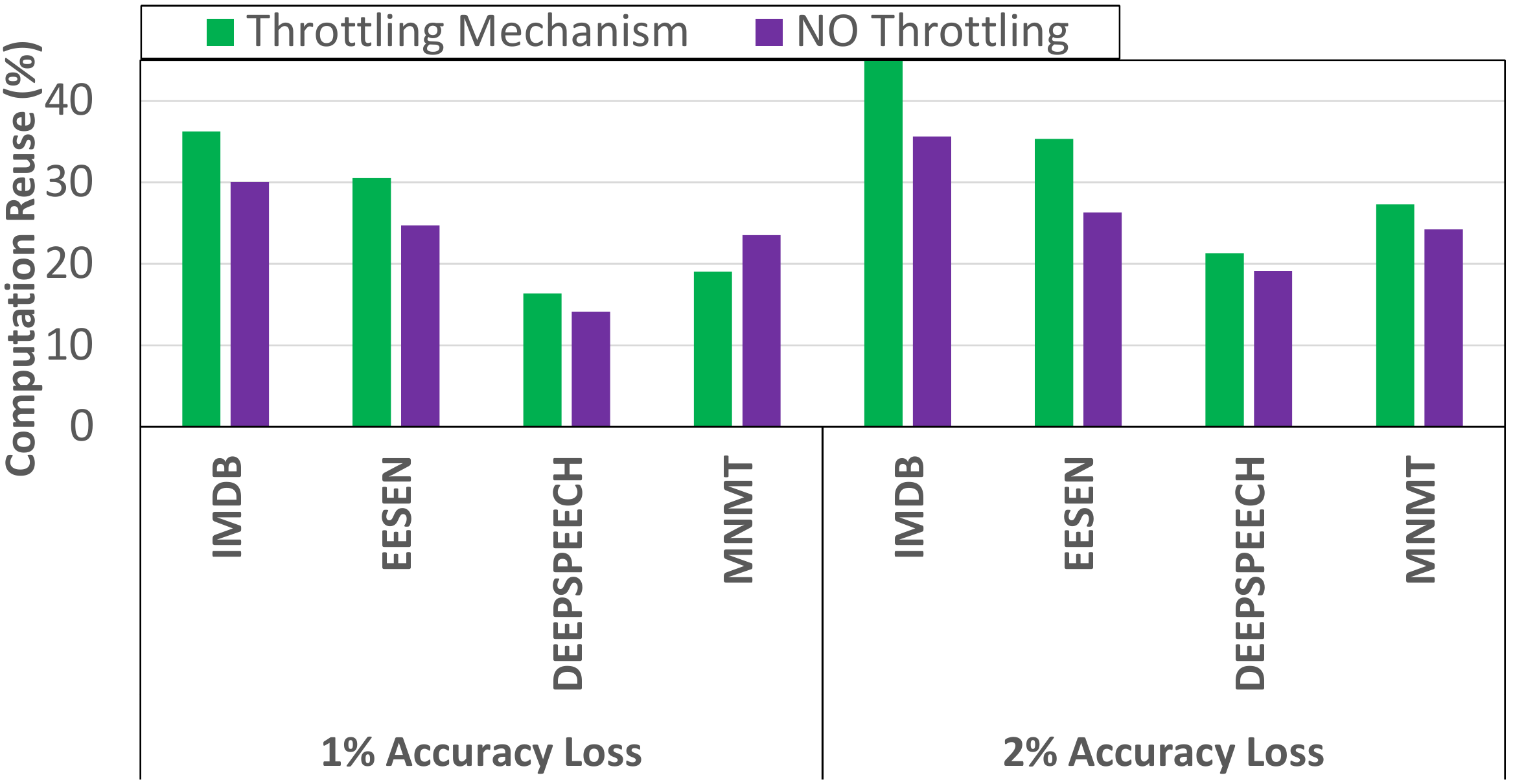}
	\caption{Computation reuse achieved by our BNN-based memoization scheme with and without the
		throttling mechanism, for accuracy losses of 1\% and 2\%. The throttling 
		mechanism provides an extra 5\% computation reuse on average for the same accuracy.}
	\label{f:rfactor_distance}
\end{figure}

We have observed that applying memoization to the same neuron in a large number
of successive time-steps may negatively impact 
accuracy, even though the relative difference $\epsilon_t^b$ in each 
individual time-step is small. We found that using a simple throttling 
mechanism can avoid this problem. More specifically,
we accumulate the relative differences over successive time-steps
where memoizaiton is applied, as shown in Equation~\ref{e:bin_delta_error}.
We use the summation of relative differences, $\delta^b_t$, to
decide whether the memoized value is reused. As illustrated in
Equation~\ref{e:yt_bin}, the memoized value is only reused when 
$\delta^b_t$ is smaller or equal than a threshold $\theta$.
Otherwise, the full-precision neuron is computed.
This throttling mechanism avoids long sequences of 
time-steps where memoization is applied to the same
neuron, since $\delta^b_t$ includes the differences
accumulated in the entire sequence of reuses.
Figure~\ref{f:rfactor_distance} shows that the 
throttling mechanism provides higher computation
reuse for the same accuracy loss.

\begin{figure}[t!]
	\centering
	\includegraphics[width=3.375in]{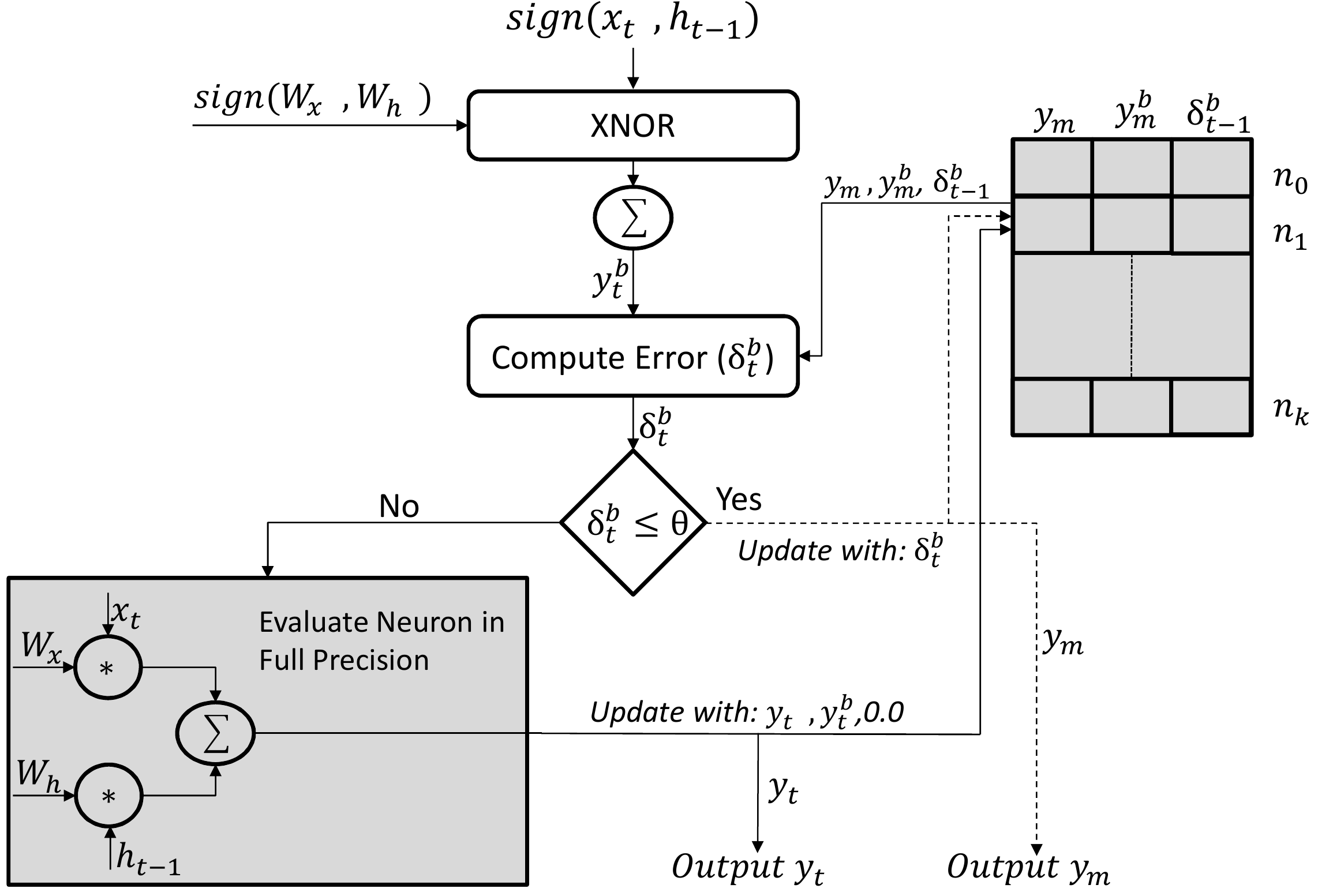}
	\caption{Fuzzy memoization scheme. $W_x$ and $W_h$ are the weights for the forward ($x_t$) 
		and recurrent connections ($h_{t-1}$) respectively. $y_t$, $y_m$ correspond to the current and 
		cached neuron output computed in full precision. $y^b_t$, $y^b_m$ are the current and cached 
		output computed by the Binary Network. $\delta_t^b$ is the summation of relative differences 
		in successive time-steps.}
	\label{f:memoization_scheme}
\end{figure}

Figure~\ref{f:memoization_scheme} summarizes the overall memoization scheme, that 
is applied to the gates in an RNN cell as follows. For the first input element ($x_0$),
i.e. the first time-step, the output values $y_0^b$ (binarized version) and $y_0$ 
(in full-precision) are computed for each neuron and stored in a memoization buffer.
$\delta^b_0$ is set to zero. In the next time-step, with input $x_1$, the value $y_1^b$ 
is computed first by the BNN. Then, the relative error ($\epsilon^b_1$) between $y_1^b$ 
and the previously cached value, $y_0^b$, is computed and added to $\delta^b_0$ to 
obtain $\delta^b_1$. Then, $\delta^b_1$ is compared with a threshold $\theta$. If  
$\delta^b_1$ is smaller than $\theta$, the cached value $y_0$ is reused, 
i.e. $y_1$ is assumed to be equal to $y_0$, and $\delta^b_1$ is stored in the 
memoization buffer. On the contrary, if $\delta^b_1$ is larger than $\theta$, the 
full precision neuron output $y_1$ is computed and its value is cached in a 
memoization buffer. In addition, $y_1^b$ is also cached and $\delta^b_1$ is set 
to zero. This process is repeated for the remaining time-steps and for all the 
neurons in each gate.  

\subsubsection{Improving the BNN Predictor Accuracy}\label{s:improving_bnn}

As discussed later in Section~\ref{s:results}, the percentage of computation reuse achieved by the BNN predictor is smaller than the oracle's percentage. Aiming to improve the BNN predictor's accuracy, we include the memoization scheme described in Section~\ref{s:memoizing_overview} during the training. The intuition is that by allowing the network to reuse similar weights (i.e., less than $\theta$) during the training,  we could transfer the obtained knowledge to the inference phase. We show in Section~\ref{s:results} that by doing this, the accuracy of the BNN predictor increases.

To include our memoization scheme into the training, we modified the forward pass as follows. First, at time-step ($t_0$), for a given neuron (i.e., $n_k$), its floating-point ($y_0$) and binarized output values $y^b_0$ are computed and cached. Second, in the next time-step ($t_1$), to set the output value of $n_k$, we first evaluate $n_k$ using its current weights and inputs. Then, we compare its binarized output value $y_1$ with its binarized output in the previous time-step $y^b_0$. If the similarity between these two values is below a threshold (i.e., $theta$), the previous output ($y_0$) is reused. Otherwise, the output value $y_1$ is cached and set as output. Finally, this process is repeated for all the time-steps and neurons in the model. Hence, our neuron-level memoization scheme is included in the inference pass during training, whereas the backward pass and update of weights are performed as usual.

Regarding the training hyper-parameters, we use the same values as the baseline implementation of the models (i.e., model without memoization). However, we train each model for several values of $thetha$ and choose the model with the highest amount of computation reuse and an accuracy equal to the baseline model.

\subsubsection{Finding the threshold value}

 The threshold $\theta$ is one of the key parameters in our scheme, and to find its value for a target accuracy loss and a given RNN model, we perform an exploration of it for different values. Each RNN model is evaluated using the training set during this process, and then the accuracy and degree of computation reuse for each threshold value is obtained. Then, for each RNN model, we select the value of $\theta$ that achieves the highest computation reuse for the target accuracy loss (i.e., less than 1\%). Note that this is done only once for each RNN model. Also, once $\theta$ is determined, it is used for inference on the test dataset.

\subsection{Hardware Implementation}\label{s:hardware_implementation}

\begin{figure}[t!]
	\centering
	\includegraphics[width=3.375in]{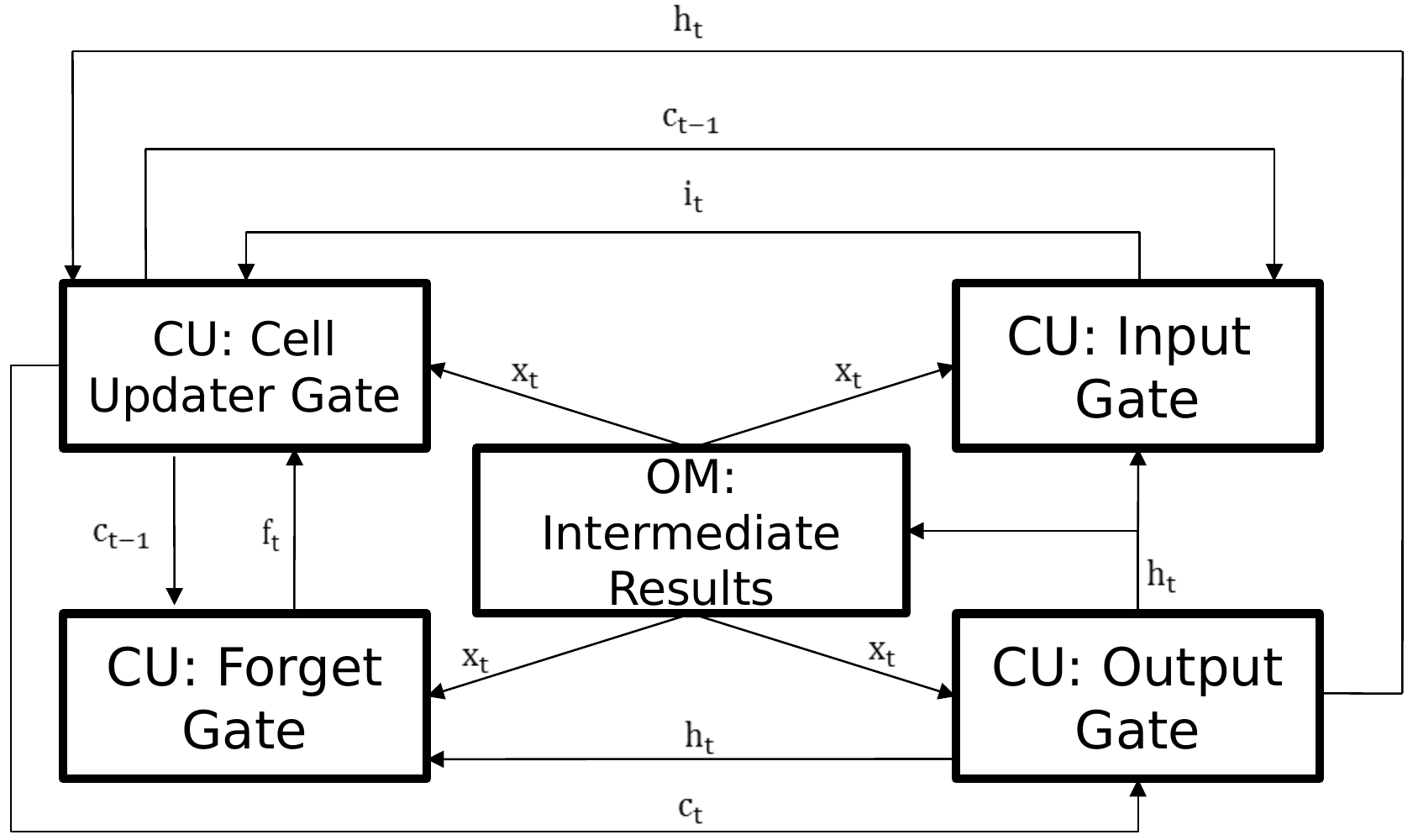}
	\caption{Overview of E-PUR architecture which consist of 4 Computation Units (CU) and an on-chip memory (OM).}
	\label{f:epur_overview}
\end{figure}

We implement the proposed memoization scheme on top of EPUR, a state-of-the-art RNN accelerator for low-power mobile applications~\cite{silfa2017epur}. 
Figure~\ref{f:epur_overview} shows a high-level block diagram of this accelerator. E-PUR is composed of four computational units tailored to the evaluation of each gate in an RNN cell and a dedicated on-chip memory used to store intermediate results. In the following subsections, we outline the E-PUR architecture's main components and detail the necessary hardware modifications required to support our fuzzy memoization scheme.

\subsubsection{Hardware Baseline}\label{s:hardware_baseline}

\begin{figure}[t!]
	\centering
	\includegraphics[width=3.375in]{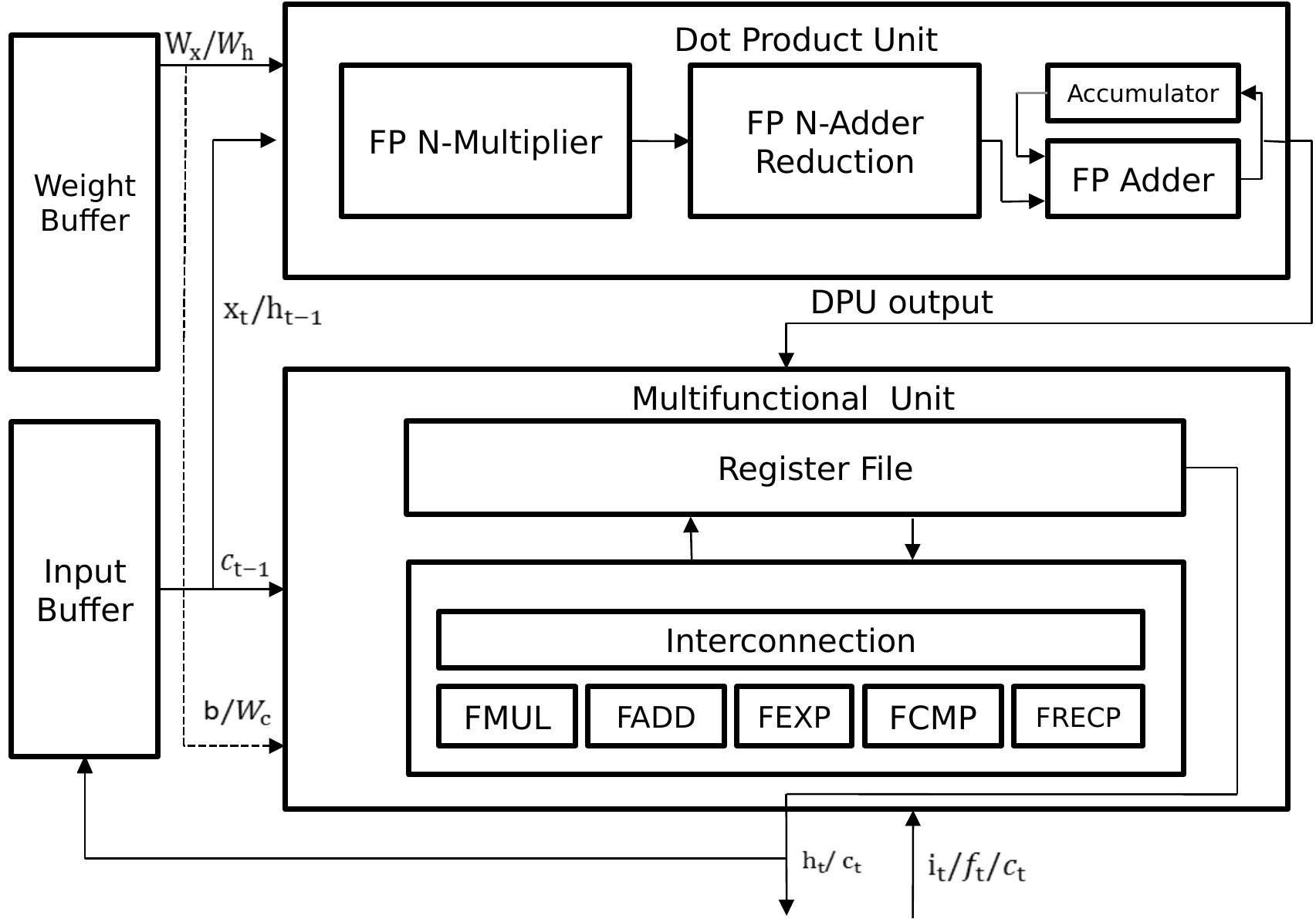}
	\caption{ Structure of E-PUR Computation Unit.}
	\label{f:computation_unit}
\end{figure}

In E-PUR each of the Computation Units (CUs), shown in Figure~\ref{f:computation_unit}, 
are composed of a dot product unit (DPU), a Multi-functional Unit (MU) and buffers to 
store the weights and inputs. The DPU is used to evaluate the matrix vector multiplications 
between the weights and inputs (i.e. $x_t$ and $h_{t-1}$) whereas the MU is used to compute 
activation functions and scalar operations. Note that in E-PUR computations can be performed using 
32 or 16 bits floating points operations.

In E-PUR, while evaluating an RNN cell, all the gates are computed in parallel for 
each input element. On the contrary, the neurons in each gate are evaluated 
in a sequential manner for the forward and recurrent connections. The following steps are executed 
in order to compute  the output value ($y_t$) for a given neuron (i.e. $n_i$). First, the input 
and weight vectors formed by the recurrent and forward connections (i.e, $x_t$ and  $h_{t-1}$) are 
split into \textit{K} sub-vectors of size \textit{N}. Then, two sub-vectors of size \textit{N} are 
loaded from the input and weight buffer respectively and the dot product between them is computed 
by the DPU, which also accumulates the result. Next, the steps are repeated for the next $k^{th}$ 
sub-vector and its result is added to the previously accumulated dot products. This process is 
repeated until all \textit{K} sub-vectors are evaluated and added together. Once the output 
value $y_t$ is computed, the DPU sends it to the MU where bias and peephole calculations are 
performed. Finally, the MU computes the activation function and stores the result in the 
on-chip memory for intermediate results. Note that once the DPU sends a value to the MU, 
it will continue with the evaluation of the next neuron output, hence, overlapping the computations
 executed by the MU and DPU since they are independent. Finally, these steps are repeated until 
all the neurons in the gate (for all cells) are evaluated for the current input element.

\subsubsection{Support for Fuzzy Memoization }\label{s:binary_memoizing_support}

In order to perform fuzzy memoization through a BNN, two modifications are done to each
 CU in E-PUR. First, the weight buffer is split into two buffers: one buffer is used to store 
the weight signs (sign buffer) and the other is used to store the remaining bits of the weights. 
Note that the sign buffer is always accessed to compute the output of the binary network ($y_t^b$) 
whereas the remaining bits are only accessed if the memoized value ($y_m$) is not reused. The binarized weights are stored in a small memory which has low energy cost but, as a consequence of splitting the weight buffer, its area  increases a bit (less than one percent).    

\begin{figure}[t!]
	\centering
	\includegraphics[width=3.375in]{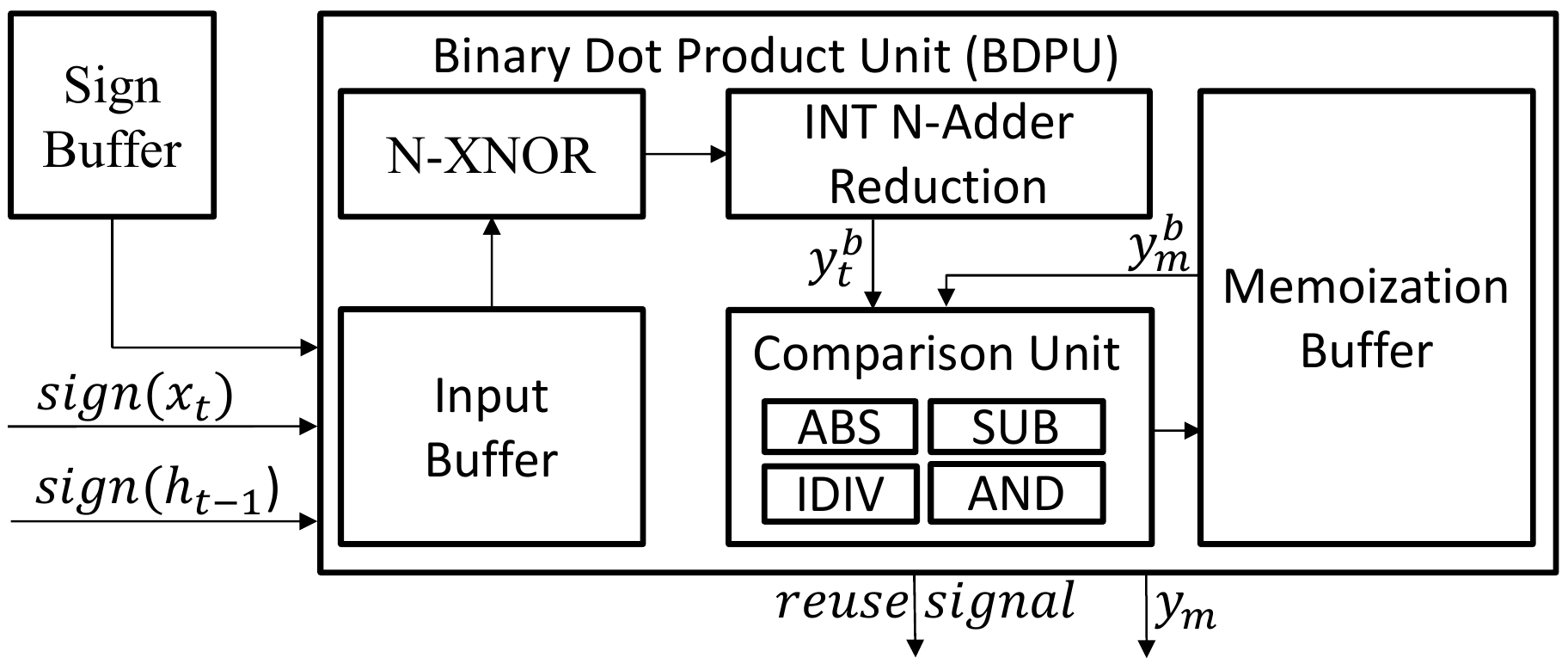}
	\caption{Structure of the Fuzzy Memoization Unit (FMU).}
	\label{f:memoizing_unit}
\end{figure}

The second modification to the CUs is the addition of the fuzzy memoization unit (FMU) which 
is used to evaluate the binary network and to perform fuzzy memoization. This unit takes as input 
two size-\textit{T} vectors (i.e., number of neurons in an RNN cell). The first vector is a weight
 vector loaded from the sign buffer whereas the other is created as the concatenation of the 
forward ($x_t$) and the recurrent connections ($h_{t-1}$).

As shown in Figure \ref{f:memoizing_unit}, the main components of the FMU are the BDPU that computes the binary dot product and the comparison unit (CMP) which decides when to 
reuse a memoized value. In addition, the FMU includes a buffer (memoization buffer) which
stores the $\delta_t^b$ for every neuron and the latest evaluation of the neurons by the full precision and binary
networks. BNN neurons (i.e, binary dot product) are evaluated using a bitwise XNOR operation and an 
adder reduction tree to gather the resulting bit vector. In the CMP unit, the relative error ($\delta_t^b$) 
is computed using integer and fixed-point arithmetic.

The steps to 
evaluate the RNN cell, described in Section \ref{s:hardware_baseline}, are executed in a 
slightly different manner to include the fuzzy memoization scheme. First, the binarized input and weight vectors for a given neuron in 
a gate are loaded into an FMU from the 
input and sign buffers respectively. Next, the BDPU computes the dot product and sends the 
result ($y_t^b$) to the comparison unit (CMP). Then, the CMP loads the previously cached 
values $y_m^b$ and $\delta^b_{t-1}$ from the memoization buffer and it uses them to compute the 
 relative error ($\epsilon^b_t$) and the $\delta^b_t$. Once $\delta^b_t$ is computed, it is compared with 
a threshold ($\theta$) to determine whether the full-precision neuron needs to be 
evaluated or the previously cached value is reused instead. In the case 
that $\delta^b_t$ is greater than $\theta$, an evaluation in full-precision is triggered. In that regard, the DPU is signaled to start the full precision evaluation which is done following the steps described in 
Section \ref{s:hardware_baseline}. After the full precision evaluation, the values $y_t$, $y_t^b$, and $0.0$ are cached in the memoization table corresponding to $y_m$, $y_m^b$, and $\delta^b_t$ respectively. On the other hand, if memoization can be applied (i.e. $\delta^b_t$ is smaller than the maximum 
allowed error), $\delta^b_t$ is updated in the memoization table and 
the memoized value ($y_m$) is sent directly to the MU (bypassing the DPU), so the full-precision evaluation of the neuron is avoided.
 Finally, 
these steps are repeated until all the neurons in a gate are evaluated for the current input element. Since LSTM or GRU gates are processed by independent CUs, the above process is executed concurrently by all gates.

\begin{table*}[t!]
	\caption{RNN Networks used for the experiments.}
	\label{t:lstm_networks}
	\centering
	\scalebox{0.76}{
	\begin{tabular}{cccccccc}
		\cellcolor[gray]{0.9}\small\textbf{Network}&\cellcolor[gray]{0.9}\small\textbf{App Domain}&\cellcolor[gray]{0.9}\small\textbf{Cell Type}&\cellcolor[gray]{0.9}\small\textbf{Layers}&\cellcolor[gray]{0.9}\small\textbf{Neurons}&\cellcolor[gray]{0.9}\small\textbf{Base Accuracy}&\cellcolor[gray]{0.9}\small\textbf{ Reuse }&\cellcolor[gray]{0.9}\small\textbf{ Dataset }  \\
		\small IMDB Sentiment~\cite{daiL15a}&\small Sentiment Classification&\small LSTM&\small 1&\small 128&\small86.5\%&\small 36.2\% &\small IMDB dataset \\
		\cellcolor[gray]{0.9}\small DeepSpeech2~\cite{deepspeech2}&\cellcolor[gray]{0.9}\small Speech Recognition&\cellcolor[gray]{0.9}\small GRU&\cellcolor[gray]{0.9}\small 5&\cellcolor[gray]{0.9}\small 800&\cellcolor[gray]{0.9}\small 10.24 WER&\cellcolor[gray]{0.9}\small 16.4\%&\cellcolor[gray]{0.9}\small LibriSpeech\\
		
		\small EESEN~\cite{miao2015eesen}&\small Speech Recognition&\small BiLSTM&\small 10&\small 320&\small 23.8     WER&\small 30.5\%&\small Tedlium V1\\
		\cellcolor[gray]{0.9}\small MNMT~\cite{britzGLL17}&\cellcolor[gray]{0.9}\small Machine Translation&\cellcolor[gray]{0.9}\small LSTM&\cellcolor[gray]{0.9}\small 8&\cellcolor[gray]{0.9}\small 1024&\cellcolor[gray]{0.9}\small29.8 Bleu&\cellcolor[gray]{0.9}\small 19.0\%&\cellcolor[gray]{0.9}\small WMT'15 En $\rightarrow$ Ge \\
	\end{tabular}
	}
\end{table*}

\begin{table}[t!]
	\caption{Configuration Parameters.}
	\label{t:epur_params}
	\centering
	\begin{tabular}{cc}
		\hline
		\multicolumn{2}{c}{\textbf{E-PUR }}\\
		\cellcolor[gray]{0.9}\small\textbf{Parameter}&\cellcolor[gray]{0.9}\small\textbf{Value}\\
		\small Technology&\small 28 nm\\
		\cellcolor[gray]{0.9}\small Frequency&\cellcolor[gray]{0.9}\small 500 MHz\\
		\small Intermediate Memory&\small 6 MiB\\
		\cellcolor[gray]{0.9}\small Weight Buffer&\cellcolor[gray]{0.9}\small 2 MiB per CU\\
		\small Input Buffer&\small 8 KiB per CU\\
		\cellcolor[gray]{0.9}\small DPU Width&\cellcolor[gray]{0.9}\small 16 operations\\
		\hline
		\multicolumn{2}{c}{\cellcolor[gray]{0.9}\textbf{Memoization Unit }}\\
		
		\small  BDPU Width&\small 2048 bits\\
		\cellcolor[gray]{0.9}\small Latency &\cellcolor[gray]{0.9}\small 5 cycles\\
		\small Integer Width&\small 2 bytes\\
		\cellcolor[gray]{0.9}\small Memoization Buffer&\cellcolor[gray]{0.9}\small 8 KiB \\
		\hline
		
	\end{tabular}
\end{table}

\section{Evaluation Methodology}\label{s:methodology}
We use a cycle-level simulator of
E-PUR customized to model our scheme as described in Section~\ref{s:binary_memoizing_support}.
This simulator estimates the total energy consumption (static and dynamic) and execution time of
the LSTM networks. The different pipeline components were 
implemented in Verilog and we synthesized them using the Synopsys Design Compiler to obtain 
their delay and energy consumption. Furthermore, we used a typical process corner with  
voltage of 0.78V. We employed CACTI~\cite{muralimanohar2009cacti} to estimate the delay
and energy consumption (static and dynamic) of on-chip memories. Finally, to estimate timing 
and energy consumption of main memory we used MICRON's memory model~\cite{Micron}. We model 4 GB of LPDDR4 DRAM.  

In order to set the clock frequency, the delays reported by Synopsys
Design Compiler and CACTI are used. We set a clock frequency that allows most hardware 
structures to operate at one clock cycle. In addition, we evaluated alternative frequency 
values in order to minimize energy consumption.

Regarding the memoization unit, the configuration parameters are shown in 
Table~\ref{t:epur_params}. Since E-PUR supports large 
LSTM networks, the memoization unit is designed to match the largest models 
supported by E-PUR. This unit has a latency of 5 clock cycles for the largest 
supported LSTM networks. In this unit, integer and fixed-point operations are 
used to perform most computations. The memoization buffer is modeled as 8KiB scratch-pad eDRAM.

The remaining configuration parameters of the accelerator used in our experiments are shown 
in Table~\ref{t:epur_params}. We strive 
to select an energy-efficient configuration for all the neural networks in Table~\ref{t:lstm_networks}.
Because the baseline accelerator is designed to accommodate large LSTM networks,
some of its on-chip storage and functional units might be oversized for some of our RNNs.
In this case, unused on-chip memories and functional units are power gated when not needed.

As for benchmarks, we use four modern LSTM networks which are described in 
Table~\ref{t:lstm_networks}. Our selection includes RNNs for popular application 
such as speech recognition, machine translation and image description. These networks have different number of internal layers and neurons. We include both bidirectional (EESEN) and unidirectional networks (the other three).
On the other hand, the length of the input sequence is also different for 
each RNN and it ranges from 20 to a few thousand input elements.

The software implementation of the networks was done in Tensorflow~\cite{tensorflow2016}.
We used the network models and the test set provided in~\cite{daiL15a,vinyalsTBE16,miao2015eesen,britzGLL17} 
for each RNN. The original accuracy for each RNN is listed in Table~\ref{t:lstm_networks}, and the accuracy loss is later reported as the absolute loss with respect to this baseline accuracy.

\section{Experimental Results}\label{s:results}

\begin{figure*}[!t]
	\centering
	\includegraphics[width=6.5in]{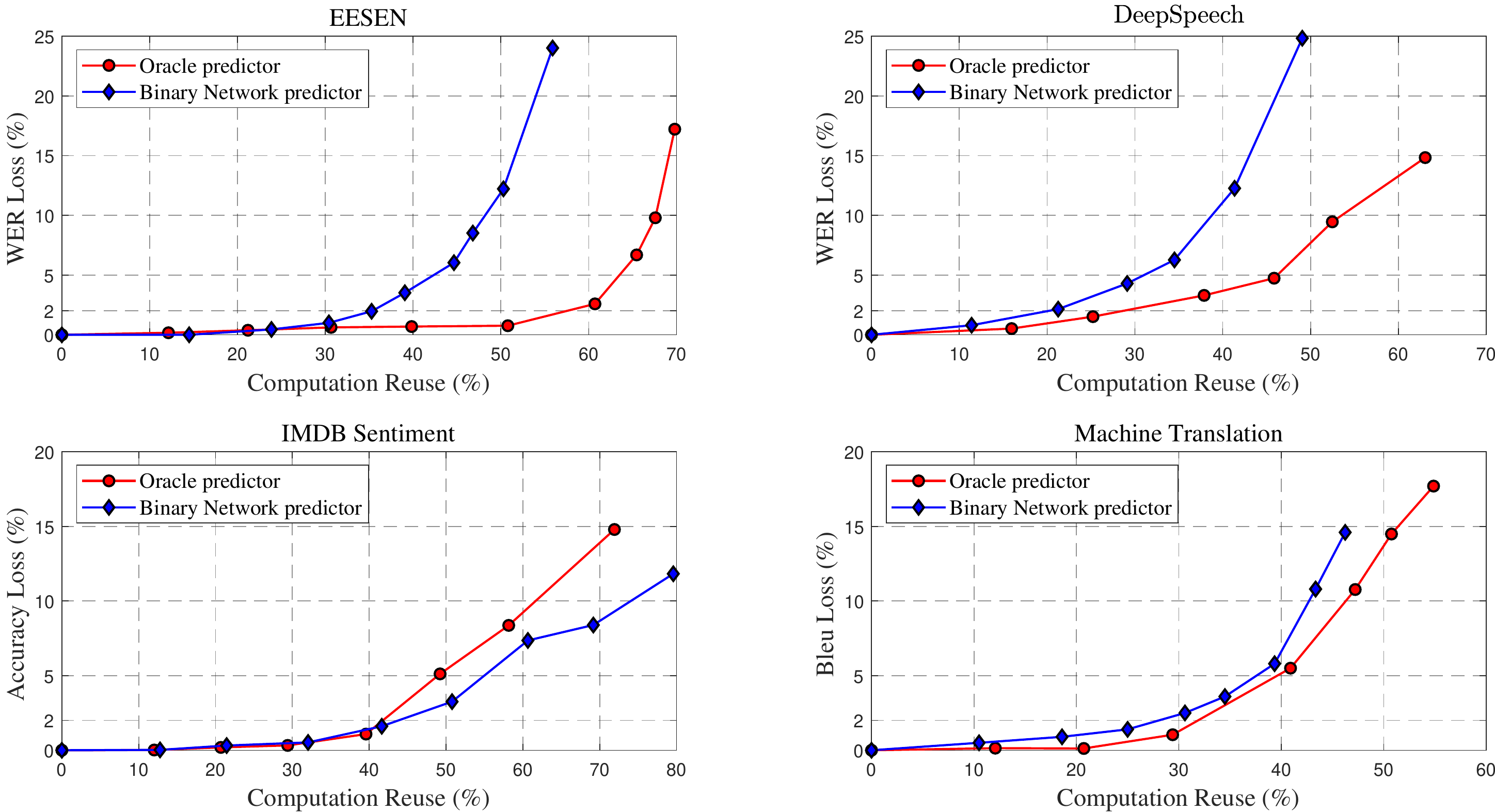}
	\caption{Percentage of computations that could be reused versus accuracy loss using Fuzzy Neuron Level Memoization with an Oracle and a Binary Network as predictors for several LSTM networks.}
	\label{f:reuse_accuracy}
\end{figure*}

This section presents the evaluation of the proposed fuzzy memoization technique for RNNs, implemented on top of E-PUR~\cite{silfa2017epur}. We refer to it as E-PUR+BM. First, we present the percentage of computation reuse and the accuracy achieved. Second, we show the performance and energy improvements, followed by an analysis of the area overheads of our technique.

Figure~\ref{f:reuse_accuracy} shows the percentage of computation reuse achieved by the BNN and
the Oracle predictors. The percentage of computation reuse indicates the percentage of neuron
evaluations avoided due to fuzzy memoization.
For accuracy losses smaller than 2\%, the BNN obtains a percentage
of computation reuse extremely similar to the Oracle. The networks EESEN and IMDB are highly tolerant 
to errors in neuron's outputs, thus, for these networks, our memoization scheme achieves reuse percentages of up to 40\% 
while having an accuracy loss smaller than 3\%. Note that, for classification problems, 
BNNs achieve an accuracy close to the state-of-the-art~\cite{rastegari2016xnor} and, 
hence, it is not surprising that the BNN predictor is highly accurate for approximating 
the neuron output.~In the case of the networks DeepSpeech (speech recognition) and NNMT (machine translation), the BNN predictor is also included in the training as discussed in Section~\ref{s:improving_bnn}. For DeepSpeech, the reuse percentage is up to 24\% for  
accuracy losses smaller than 2\%. In this network, the input sequence tends to be large 
(i.e, 900 elements on average). As the reuse is increased, the error introduced to 
the output sequence of a neuron persists for a larger number of elements. Therefore, 
the introduced error will have a bigger impact both in the evaluation of the current layer, due
to the recurrent connections, and the following layers. As a result, the overall accuracy 
of the network decreases faster.~For MNMT, the BNN predictor and the oracle achieve 
similar reuse versus accuracy trade-off for up to 32\% of computation reuse. Note that, 
for this network, the linear correlation between the BNN and the full precision neuron 
output is typically lower than for the other networks in the benchmark set.

\begin{figure}[t!]
	\centering
	\includegraphics[width=3.375in]{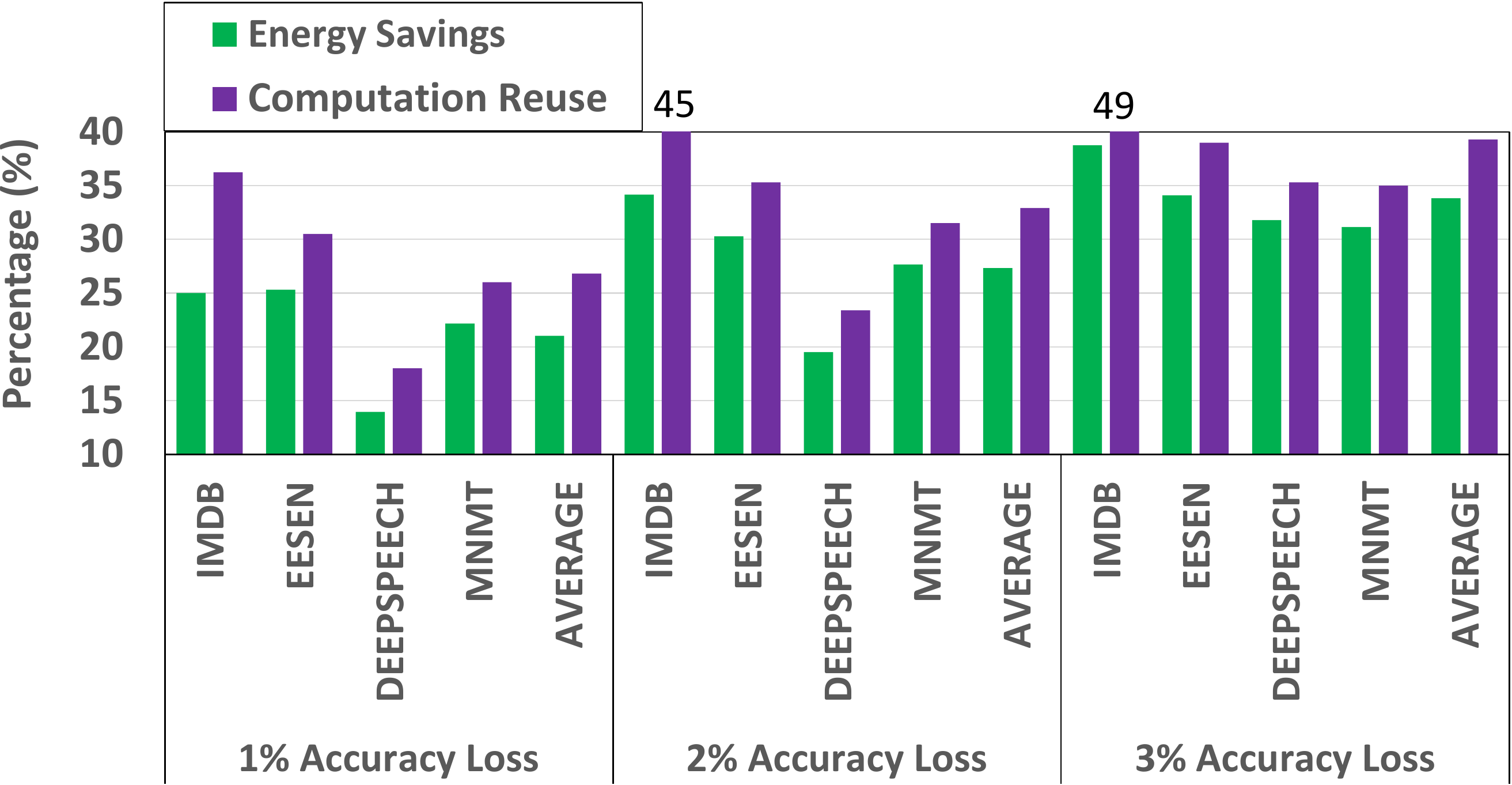}
	\caption{Energy savings and computation reuse of E-PUR+BM with respect to the baseline.}
	\label{f:energy_reuse}
\end{figure}

Figure~\ref{f:energy_reuse} shows the energy savings and computation reuse achieved
by our scheme, for different thresholds of accuracy loss.
For a conservative loss of 2\%, the average energy saving is 27.3\%, whereas the reuse 
percentage is 33\%. In this case, the networks DeepSpeech and MNMT have energy savings of 19.5\% and 27.6\%, respectively. In contrast, IMDB and EESEN are more tolerant of errors in the neuron output; thus, they exhibit the most considerable savings, 34.2\% and 30\%, respectively. For a highly conservative 1\% of accuracy loss, the computation reuse and energy saving are 26.82\% and 21\% on average, respectively. EESEN and DeepSpeech achieve 25.3\% and 14\% energy savings, respectively, for a 1\% accuracy loss. Regarding the MNMT and IMDB networks, the energy savings for 1\%  accuracy loss are 22.2\% and 25\%, respectively.

\begin{figure}[t!]
	\centering
	\includegraphics[width=3.375in]{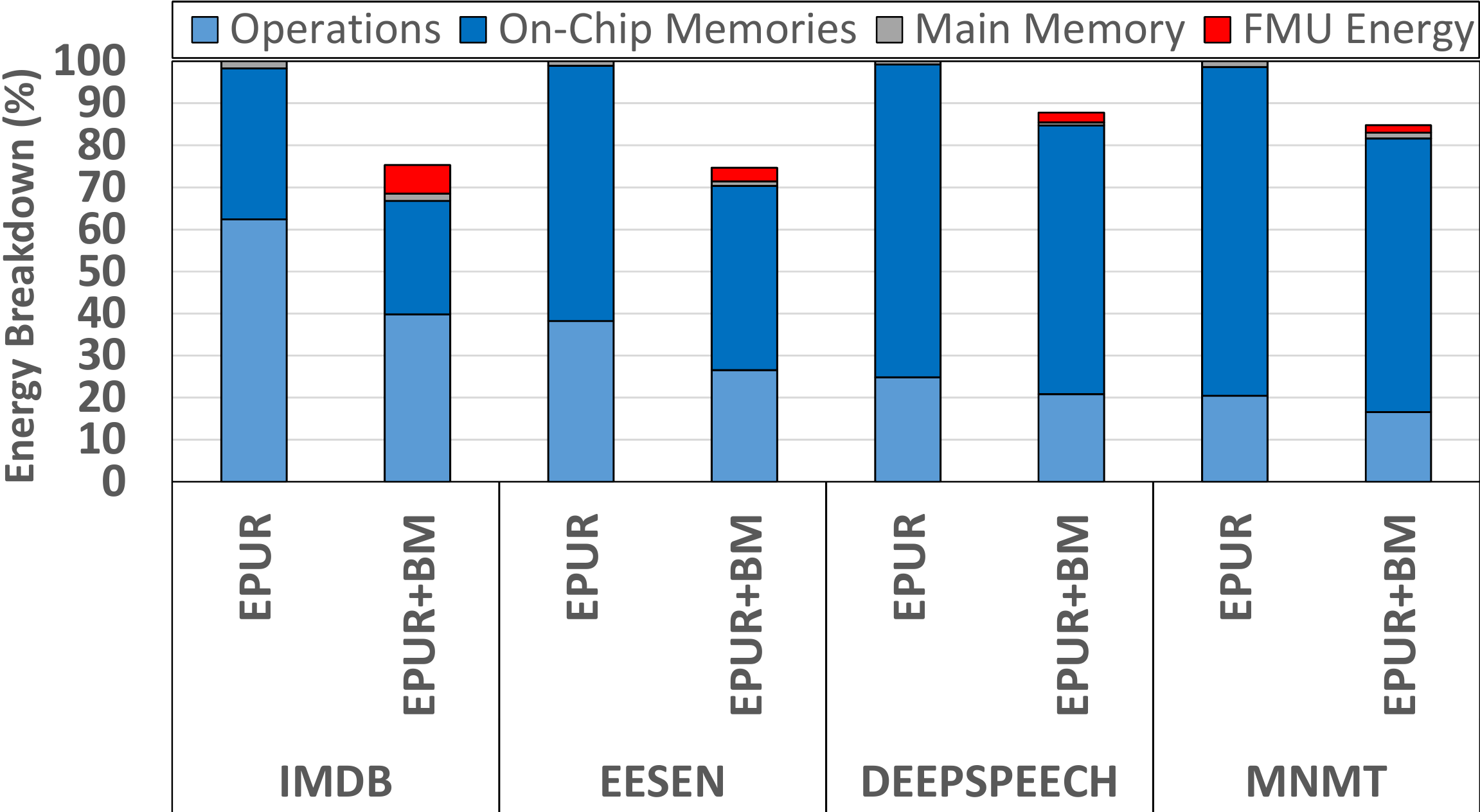}
	\caption{Energy breakdown for E-PUR and EPUR+BM. FMU Energy is the overhead due to the memoization scheme.}
	\label{f:energy_breakdown}
\end{figure}

Regarding the sources of energy savings, Figure~\ref{f:energy_breakdown} reports the energy breakdown, including static and dynamic energy, for the baseline accelerator and E-PUR+BM, for an accuracy loss of 1\%. The sources of energy consumption are grouped into on-chip memories ("scratch-pad"
memories), pipeline components ("operations", i.e. multipliers), main memory
(LPDDR4) and the energy consumed by our FMU component. Note that most of the energy consumption
is due to the scratch-pad memories and the pipeline components, and, as it can be seen,
both are reduced when using our memoization scheme. In E-PUR+BM, 
each time a value from the memoization buffer is reused, we avoid accessing all the neuron's weights and the input buffers, achieving significant energy savings. Besides, since the extra buffers used by E-PUR+BM are fairly small (i.e., 8 KB), the energy overhead due to the
memoization scheme is negligible. The energy consumption due to the operations is also reduced, as the memoization scheme avoids neuron computations.
Furthermore, the leakage of scratch-pad and operations are also reduced since the memoization scheme decreases the execution time. Finally, the energy consumption due to accessing the main memory is not affected by our technique since both E-PUR and E-PUR+BM must access the main memory to load all the weights once for each input sequence.

\begin{figure}[t!]
	\centering
	\includegraphics[width=3.375in]{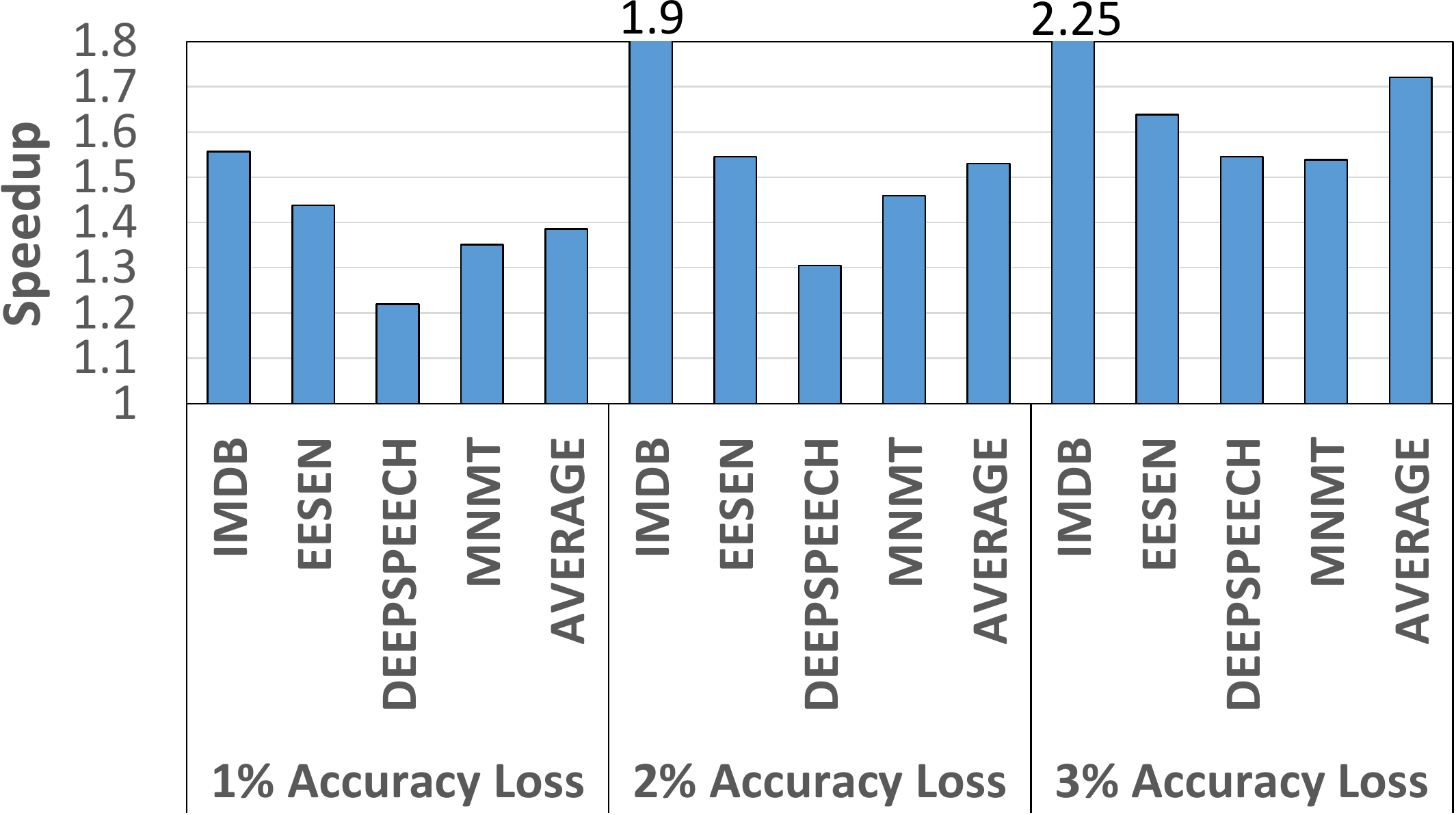}
	\caption{Speedup of E-PUR+BM over the baseline (E-PUR).}
	\label{f:time_reuse}
\end{figure}

Figure~\ref{f:time_reuse} shows the performance improvements for the different RNNs. On average, 
a speedup of 1.4x is obtained for a 1\% accuracy loss, whereas accuracy losses of 2\% and 
3\% achieve improvements of 1.5x and 1.7x, respectively. The performance improvement comes from avoiding the dot product computations for the memoized neurons. Therefore, the larger the
degree of computation reuse, the more significant the performance improvement. Note that the memoization
scheme introduces an overhead of 5 cycles per neuron (see Table~\ref{t:epur_params}) which is mainly due to the evaluation of the binarized neuron. If the full-precision neuron computation is avoided, our scheme saves between 16 and 80 cycles depending on the RNN. Therefore, 
configurations with a low degree of computation reuse, like Deepspeech at 1\% accuracy loss, 
exhibit more minor speedups due to the memoization scheme's overhead. On the other hand, 
RNNs that show higher computation reuse, such as EESEN at 2\% accuracy loss, 
achieve a speedup of 1.55x.

\begin{figure}[t!]
	\centering
	\includegraphics[width=3.375in]{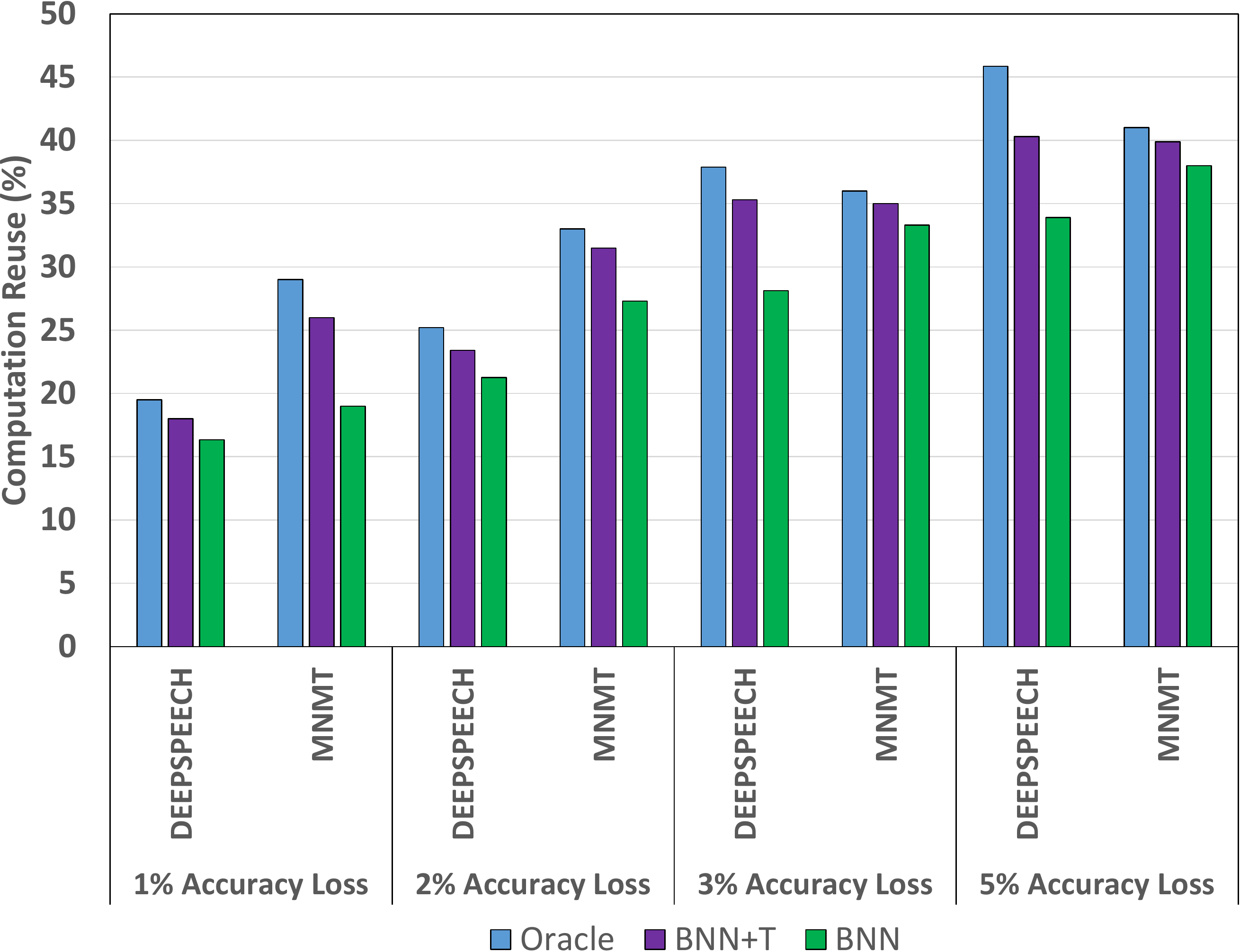}
	\caption{Computation reuse achieved by our BNN-based memoization scheme. BNN+T and BNN refer to our scheme on a model trained with and without memoization, respectively.  }
	\label{f:reuse_training}
\end{figure}

\begin{figure}[t!]
	\centering
	\includegraphics[width=3.375in]{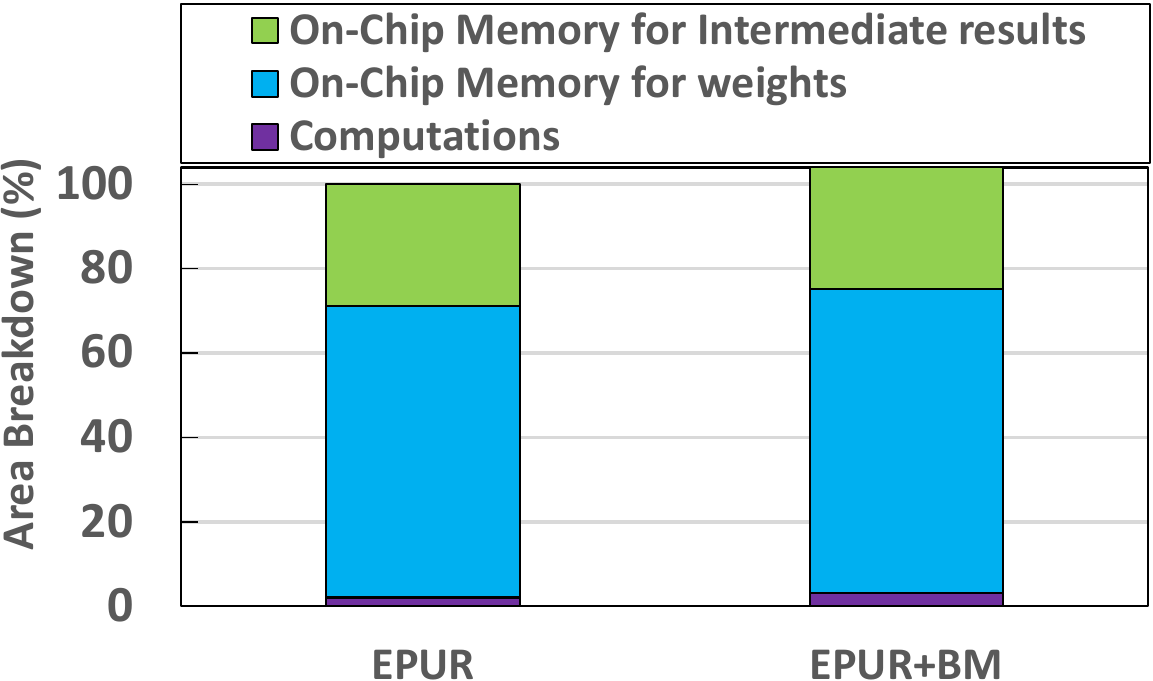}
	\caption{Area breakdown for E-PUR and EPUR+BM.  }
	\label{f:area_break}
\end{figure}

Figure~\ref{f:reuse_training} shows the accuracy and computation reuse for the oracle predictor and our memoization scheme using two different configurations.  The configuration BNN refers to the evaluation of a trained model without memoization, whereas the configuration BNN+T includes our memoization scheme on the training phase, as explained in Section 7. As shown in Figure~\ref{f:reuse_training},  for the Deepspeech model, the computation reuse is  13.9\% for an accuracy loss of 1\%. Note that the percentage of reuse increases by around 4\%, compared to the implementation that does not include the memoization scheme during training. For the NNMT model, the reuses percentages also increased by 4\% when adding our scheme to the training.

Figure~\ref{f:area_break} shows the area breakdown of E-PUR and E-PUR+BM.  
Regarding the area, E-PUR has an area of 64.6 $mm^2$, whereas E-PUR+BM requires 66.8 $mm^2$ (4\% area overhead). As shown in Figure 21, the area for the on-chip memories to store the weights is 69\% and 72\%, for E-PUR and E-PUR+BM, respectively. E+PUR+BM requires an extra 3\% since the on-chip memories for the weights are split into two separate banks: storing the BNN and the other to store the full-precision weights. The computations' area requirements are 2\% and 3\%, for E-PUR and E-PUR+BM, respectively. The overhead due to computations comes from the extra logic added to implement the memoization unit.

\section{Related Work}\label{s:related_work}

Increasing energy-efficiency and performance of LSTM networks has attracted 
the attention of the architectural community in recent 
years~\cite{Han:2017:EES:3020078.3021745,li2015fpga,guan2017fpga,lee2016fpga}.
Most of these works employ pruning and compression techniques to improve
performance and reduce energy consumption. Furthermore, linear quantization 
is employed to decrease the memory footprint. On the contrary, our technique 
improves energy-efficiency by relying solely on computation reuse at the neuron 
level. To the best of our knowledge, this is the first work using a BNN as a 
predictor for a fuzzy memoization scheme. BNNs have been used 
previously~\cite{courbariaux2016binarized,rastegari2016xnor,kim2016bitwise} 
as standalone networks, whereas we employs BNNs in conjunction
with the LSTM network to evaluate neurons on demand.

Fuzzy memoization has been  extensively researched in the past and 
has been implemented both in hardware and software. Hardware schemes to 
reuse instructions have been proposed in~\cite{sodani1997,alvarez2001PTR,gonzalezTM99,arnau2014}.
Alvarez et al.~\cite{alvarez2005FMF} presented a fuzzy memoization scheme to improve performance of floating point operations 
in multimedia applications. In their scheme floating point operations are memoized using a 
hash of the source operands, whereas in our technique, a whole function (neuron inference) is memoized 
based on the values predicted by a BNN.     

Finally, software schemes to memoize entire functions have been presented 
in the past~\cite{xu2007dpa,acar2003SM}. These schemes are tailored to general purpose 
programs whereas our scheme is solely focused in LSTM networks, since it exploits
the intrinsic error tolerance of LSTM networks.

\section{Conclusions}\label{s:conclusions}

This paper has shown that 25\% of neurons in an LSTM network change their output value by less than 10\%, which motivated us to propose a fuzzy memoization scheme to
save energy and time. A significant challenge to perform neuron-level fuzzy memoization is to predict accurately, in a simple manner, whether the output of a given neuron will be similar to a previously computed and cached value. To this end, we propose to use a 
Binarized Neural Network (BNN) as a predictor, based on the observation that the full-precision output of a neuron is highly correlated with the output of the corresponding 
BNN. We show that a BNN predictor achieves 26.7\% computation reuse on average, which is very similar to the results obtained with an Oracle predictor. Moreover, we have shown that including
our technique during the training phase further improves the BNN predictor's accuracy by 4\% or more. 
We have implemented our technique on top of E-PUR, a state-of-the-art accelerator for 
LSTM networks. Results show that our memoization scheme achieves significant time and energy savings with minimal impact on the accuracy of the RNNs. When compared 
with the E-PUR accelerator, our system achieves 21\% 
energy savings on average, while providing 1.4x speedup at the expense of a minor 
accuracy loss.

\section*{Acknowledgments}

This work has been supported by the CoCoUnit ERC Advanced Grant of the EU's Horizon 2020 program (grant No 833057), the Spanish State Research Agency (MCIN/AEI) under grant PID2020-113172RB-I00, and the ICREA Academia program.

\bibliographystyle{IEEEtranS}
\bibliography{ref}

\end{document}